\documentclass[lettersize,journal]{IEEEtran}
\usepackage{amsmath,amsfonts}
\usepackage{algorithmic}
\usepackage{algorithm}
\usepackage{array}
\usepackage{textcomp}
\usepackage{stfloats}
\usepackage{url}
\usepackage{verbatim}
\usepackage{graphicx}
\usepackage{cite}
\usepackage{multirow}
\usepackage{caption}
\usepackage[pagebackref,breaklinks,colorlinks]{hyperref}
\usepackage{makecell}
\begin{document}

\title{Global Modeling Matters: A  Fast, Lightweight and Effective Baseline for Efficient Image Restoration}

\author{Xingyu Jiang, Ning Gao, Hongkun Dou, Xiuhui Zhang, Xiaoqing Zhong, \\ Yue Deng,~\IEEEmembership{Senior Member,~IEEE,} and Hongjue Li

\thanks{Xingyu Jiang, Ning Gao, Hongkun Dou, Xiuhui Zhang, Yue Deng  and Hongjue Li are with Beihang University, Beijing 100191, China;  Xiaoqing Zhong is with the China Academy of Space Technology.}
}

\markboth{Journal of \LaTeX\ Class Files,~Vol.~14, No.~8, August~2021}%
{Shell \MakeLowercase{\textit{et al.}}: A Sample Article Using IEEEtran.cls for IEEE Journals}


\maketitle

\begin{abstract}
Natural image quality is often degraded by adverse weather conditions, significantly impairing the performance of downstream tasks. Image restoration has emerged as a core solution to this challenge and has been widely discussed in the literature. Although recent transformer-based approaches have made remarkable progress in image restoration, their increasing system complexity poses significant challenges for real-time processing, particularly in real-world deployment scenarios. To this end, most existing methods attempt to simplify the self-attention mechanism, such as by channel self-attention or state space model. However, these methods primarily focus on network architecture while neglecting the inherent characteristics of image restoration itself. In this context, we explore a pyramid Wavelet-Fourier iterative pipeline to demonstrate the potential of Wavelet-Fourier processing for image restoration. Inspired by the above findings, we propose a novel and efficient restoration baseline, named Pyramid Wavelet-Fourier Network (PW-FNet). Specifically, PW-FNet features two key design principles: 1) at the inter-block level, integrates a pyramid wavelet-based multi-input multi-output structure to achieve multi-scale and multi-frequency bands decomposition; and 2) at the intra-block level, incorporates Fourier transforms as an efficient alternative to self-attention mechanisms, effectively reducing computational complexity while preserving global modeling capability. Extensive experiments on tasks such as image deraining, raindrop removal, image super-resolution, motion deblurring, image dehazing, image desnowing and underwater/low-light enhancement demonstrate that PW-FNet not only surpasses state-of-the-art methods in restoration quality but also achieves superior efficiency, with significantly reduced parameter size, computational cost and inference time. The code is available at: https://github.com/deng-ai-lab/PW-FNet.
\end{abstract}


\section{Introduction}
\IEEEPARstart{N}{atural} image quality deteriorates significantly under adverse weather conditions, severely impacting the performance of subsequent high-level vision tasks. This degradation not only reduces visibility but also disrupts critical feature extraction, which poses challenges for tasks such as object detection \cite{object}, segmentation \cite{seg}, and tracking \cite{track}. As a result, image restoration has been proposed as a preprocessing solution aimed at restoring degraded images to their clean counterparts, drawing considerable attention from researchers. However, due to the inherently ill-posed nature of image restoration, traditional prior-based methods \cite{chen2013generalized, luo2015removingDSC} struggle to handle complex real-world scenarios effectively, as their predefined constraints often fail to generalize across diverse degradation patterns and background textures. In recent years, deep learning-based approaches \cite{zheng2022sapnetSAPNet, RESCAN, PReNet} have achieved remarkable success. Specifically, convolutional neural networks (CNNs), as a type of local perceptual network, have demonstrated strong restoration capabilities through end-to-end learning. More recently, transformer-based models \cite{Restormer, DRSformer, Uformer} have significantly advanced the field by leveraging self-attention mechanisms to incorporate global contextual information, surpassing CNN-based restoration performance in challenging conditions.

\begin{figure}[t]
	\centering
	\includegraphics[width=\linewidth]{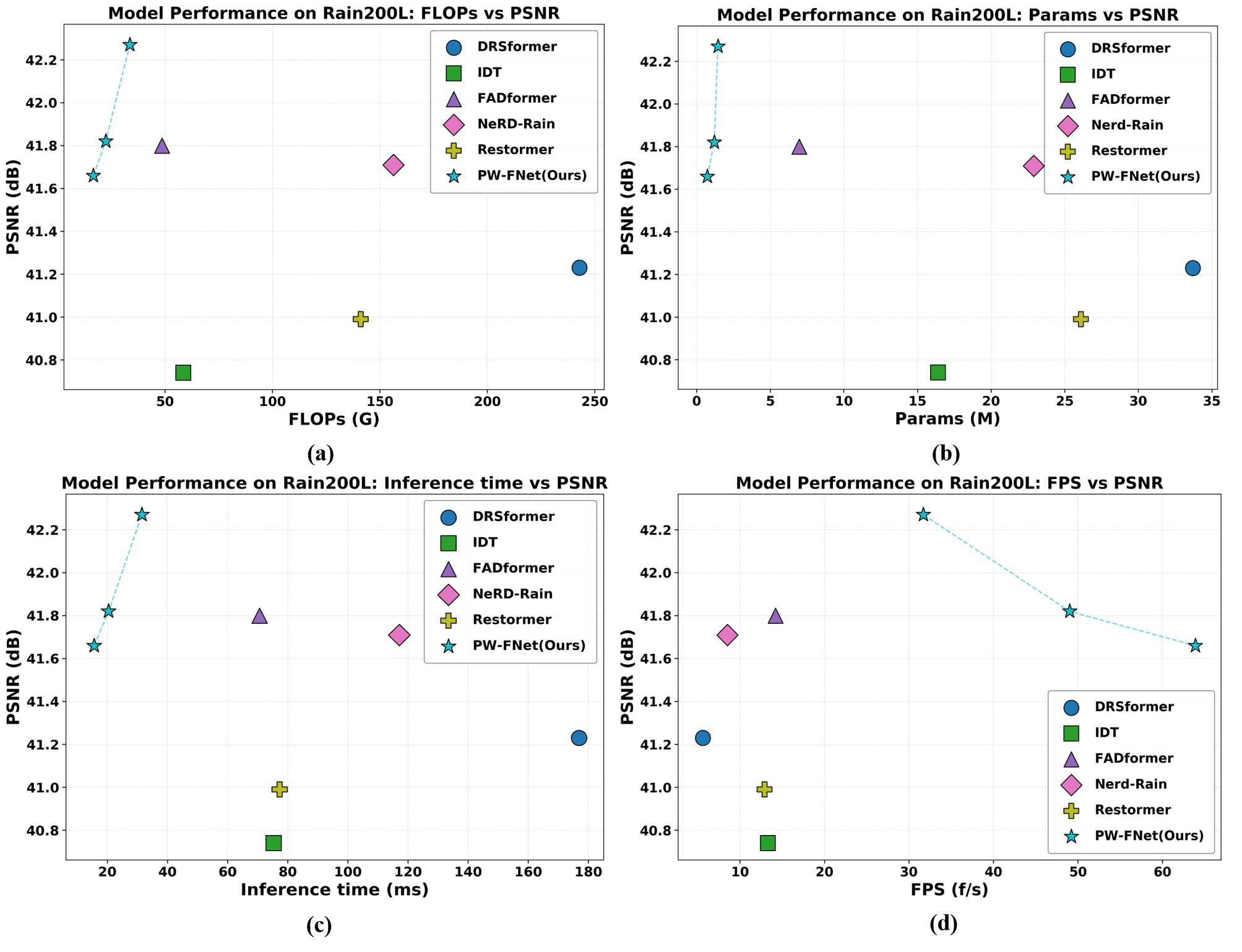}
	\caption{Comparison with different state-of-the-art approaches on the Rain200L\cite{RAIN200} dataset: (a) FLOPs vs PSNR; (b) Parameters vs PSNR; (c) Inference time vs PSNR; (d) FPS vs PSNR. Our method achieves better performance and all calculations are performed at a resolution of $256\times256$.
	}
	\label{fig:compa_intro}
    \vspace{-2mm}
\end{figure}

\begin{figure*}[t]
\setlength{\abovecaptionskip}{0cm}
\setlength{\belowcaptionskip}{-0.cm}
	\centering
	\includegraphics[width=\linewidth]{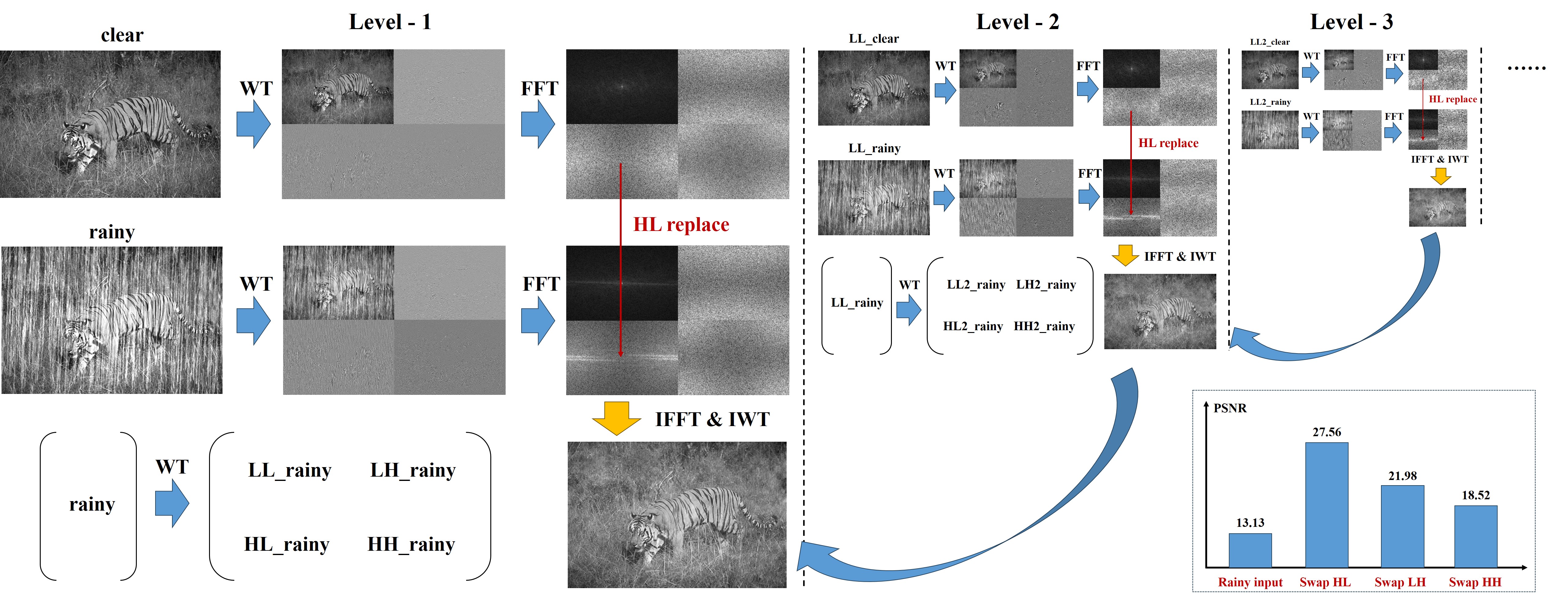}
	\caption{The pyramid wavelet-Fourier iterative pipeline for image restoration. Rain degradation is concentrated on HL sub-bands.
	}
	\label{fig:intro}
    \vspace{-3mm}
\end{figure*}

Despite above advancements, due to the quadratic computational complexity of self-attention, transformer-based models \cite{DRSformer, Uformer} require high system complexity (see Fig.\ref{fig:compa_intro}(a) and Fig.\ref{fig:compa_intro}(b)), which limits their practical applicability in real-time or resource-constrained environments (see Fig.\ref{fig:compa_intro}(c) and Fig.\ref{fig:compa_intro}(d)), such as autonomous driving \cite{auto}. Consequently, existing methods predominantly focus on simplifying self-attention mechanisms in two primary ways: \textbf{1) simplifying the self-attention mechanism itself}, where various methods aim to restructure the self-attention computation to reduce complexity. Notable examples include Restormer, which introduces channel self-attention, and GRL, which incorporates anchor self-attention; \textbf{2) replacing the self-attention mechanism}, inspired by the Metaformer concept. For instance, NAFNet replaces self-attention with convolution, while the latest Mamba-based methods substitute self-attention with state space models. We follow the latter manner, as we believe the key to the success of transformers in image restoration lies in their global modeling capability, with self-attention being merely one possible avenue for achieving this. In this context, rather than exclusively focusing on model architecture design, we investigate efficient global modeling strategies grounded in the inherent priors of the image restoration tasks themselves to develop a lightweight and effective restoration baseline.


To this end, we present the pyramid wavelet-Fourier iterative pipeline, without any deep learning techniques, to efficiently separate the groundtruth and degraded components in the Wavelet-Fourier domain, as illustrated in Fig.\ref{fig:intro}. Taking deraining as an example, in Level-1 of Fig.\ref{fig:intro}, we begin by applying wavelet transform (WT) to the input rainy and clear images, decomposing them into four sub-bands: [LL, LH, HL, HH]. Subsequently, we apply fast Fourier transform (FFT) to each sub-band. The Fourier transform's global operation reveals that degradation primarily concentrates in the compact high-frequency regions of the [LL, HL] sub-bands. In Level-2, we take the LL sub-band from Level-1 as input and repeat the process, observing a similar pattern: the degradation differences remain localized in the high-frequency regions of the [LL, HL] sub-bands. This iterative decomposition continues until the resolution of the LL sub-band becomes sufficiently small, at which point we swap the HL sub-band between the rainy and clear images. The derained image is then reconstructed through inverse Fourier and inverse wavelet transforms. Additionally, we present the quantitative results of swapping different sub-bands in the bottom-right corner of Fig.\ref{fig:intro}. This wavelet-Fourier pipeline highlights two key observations: \textbf{1) wavelet iterative decomposition effectively isolates degradation components into specific sub-band, thereby reducing the feature processing space; 2) Fourier transform’s global operation further concentrates degradation into compact high-frequency regions, narrowing the feature extraction scope.} By precisely localizing degradation-related features, the Wavelet-Fourier pipeline demonstrates its potential for achieving effective global restoration capabilities while maintaining computational efficiency.

Building upon the aforementioned findings, we propose a powerful and efficient restoration baseline, the Pyramid Wavelet-Fourier Network (PW-FNet), which integrates pyramid wavelet-Fourier transforms into the model architecture to achieve efficient image restoration. Specifically, at the inter-block level, we incorporate a pyramid wavelet multi-input multi-output structure, enabling multi-scale and multi-frequency bands decomposition, while allowing for dynamic adaptive output. This design enables a single trained model to generate small, medium and large versions without additional training. At the intra-block level, we replace self-attention mechanisms with Fourier transforms, enabling efficient global feature modeling while significantly reducing network complexity. To validate the effectiveness of PW-FNet, we conduct extensive experiments on a range of image restoration tasks, including deraining, raindrop removal, super-resolution, motion deblurring, dehazing, desnowing, and underwater/low-light enhancement and demonstrate that PW-FNet not only outperforms state-of-the-art methods in restoration quality but also achieves superior efficiency, with significantly reduced parameter size, computational cost and inference time.



We summarize the mian contributions as three-folds:
\begin{itemize}
\item[$\bullet$] We show the pyramid Wavelet-Fourier iterative pipeline, which effectively decomposes degradation into distinct sub-bands through wavelet transforms and isolates compact feature regions using Fourier transforms.
\item[$\bullet$] We propose a powerful and efficient restoration baseline, which replaces self-attention mechanisms with pyramid Wavelet-Fourier transforms to achieve efficient multi-scale and multi-frequency bands global modeling.
\item[$\bullet$] We thoroughly validate PW-FNet across deraining, raindrop removal, super-resolution, motion deblurring, dehazing, desnowing and underwater/low-light enhancement, demonstrating its superior performance with lightweight parameter size, computational cost and inference time.
\end{itemize}

\begin{figure*}[t]
	\centering
	\includegraphics[width=\linewidth]{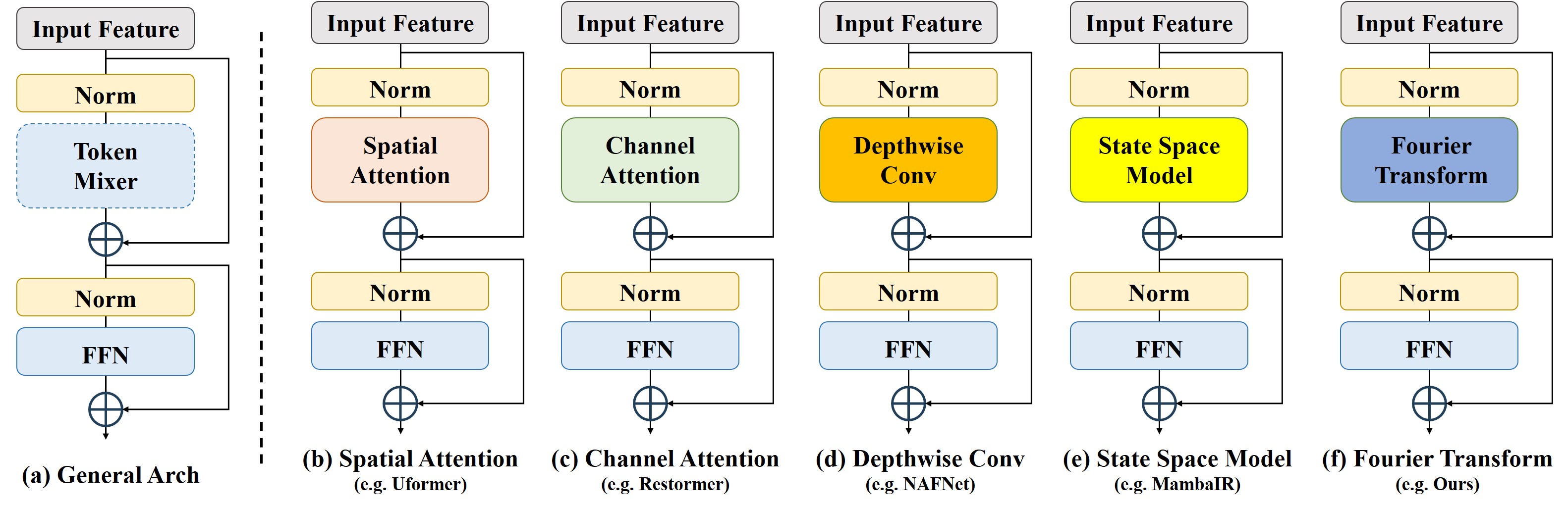}
	\caption{Comparison of intra-block architecture designs for global modeling across various restoration baselines.
	}
	\label{fig:mixer}
    \vspace{-2mm}
\end{figure*}

\section{Related Works}

\subsection{Image Restoration}

In recent years, deep learning has driven significant advancements in image restoration. Early CNN-based methods (e.g., DDN \cite{fu2017removingDDN}) framed image restoration as an image-to-image translation task. Subsequently, models such as RESCAN \cite{li2018recurrentRESCAN}, PReNet \cite{ren2019progressivePReNet}, and MSPFN \cite{jiang2020multiMSPFN} introduced recurrent structures and multi-scale architectures to enhance restoration performance. However, CNNs face challenges in long-range global modeling, limiting their effectiveness in handling complex degradation scenarios. To address this, transformer-based approaches (e.g., IDT \cite{xiao2022imageIDT}, Uformer \cite{wang2022uformerUformer}) leverage self-attention mechanisms to capture global context, achieving superior restoration performance. Despite this, their high computational cost impedes real-time applications, prompting the exploration of lightweight self-attention variants and alternative paradigms.

As illustrated in Fig.\ref{fig:mixer}, we summarize the intra-block design of various baseline architectures in the image restoration field. Generally, most existing methods adhere to the standard architecture shown in Fig.\ref{fig:mixer}(a), which consists of a Token Mixer followed by a Feed Forward Network (FFN). Various works focus on designing different token mixers within this architecture to enhance modeling capabilities. For instance, Uformer \cite{Uformer} employs Swin Transformer \cite{liu2021swin}’s window attention to achieve spatial global modeling. Restormer \cite{zamir2022restormerRestormer} innovatively shifts self-attention to the channel dimension to reduce computational complexity. More recently, inspired by MetaFormer \cite{yu2022metaformer}, models such as NAFNet \cite{NAFNet} and Mambair replace attention mechanisms with depthwise convolutions and state space models respectively, achieving competitive performance with less model complexity. Distinct from these methods, we replaces self-attention mechanisms with pyramid Wavelet-Fourier transforms to achieve efficient multi-scale and multi-frequency bands global modeling.

\subsection{Wavelet and Fourier Processing}

In the field of image restoration, both wavelet and Fourier domains have been extensively explored and applied to tackle various degradation challenges. Wavelet transforms, in particular, excel in extracting multi-scale and multi-frequency features. Many methods leverage wavelet techniques to decompose images into different frequency bands, facilitating effective feature separation. For example, DAWN\cite{dawn} introduces vector decomposition and a direction-aware attention mechanism for wavelet-based image deraining. Similarly, Wave-mamba\cite{wavemamba} combines wavelet transforms with state space models to achieve lossless low-light enhancement.

Fourier transforms, on the other hand, are highly effective in capturing global frequency characteristics and have been successfully applied to a variety of image restoration tasks. The main challenge here lies in differentiating the learning of various frequency components, particularly given the imbalance between low-frequency and high-frequency information. For instance, SFNet\cite{sfnet} designs learnable convolutions for both low and high-frequency components to perform frequency selection. SFHformer\cite{jiang2024SFHformer} adopts frequency-dynamic convolutions to extract features from distinct frequency components, while DMSR\cite{DMSR} employs a spatial-frequency multi-branch architecture to capture more comprehensive properties.

In our work, we explore the potential of combining wavelet and Fourier transforms, harnessing the strengths of both approaches. Specifically, we begin by decomposing images into different frequency sub-bands using wavelets and then refine the results through Fourier-based global modeling. This hybrid approach enables efficient degradation decomposition and compact feature extraction, making it highly effective for a wide range of image restoration tasks. In particular, our proposed Pyramid Wavelet-Fourier Network (PW-FNet) combines the advantages of both wavelet and Fourier transforms to achieve multi-scale, multi-frequency global modeling. By replacing traditional self-attention mechanisms with pyramid wavelet-Fourier transforms, PW-FNet provides an efficient solution to image restoration while maintaining efficiency.

\begin{figure*}[t]
\centering
\includegraphics[width=\linewidth]{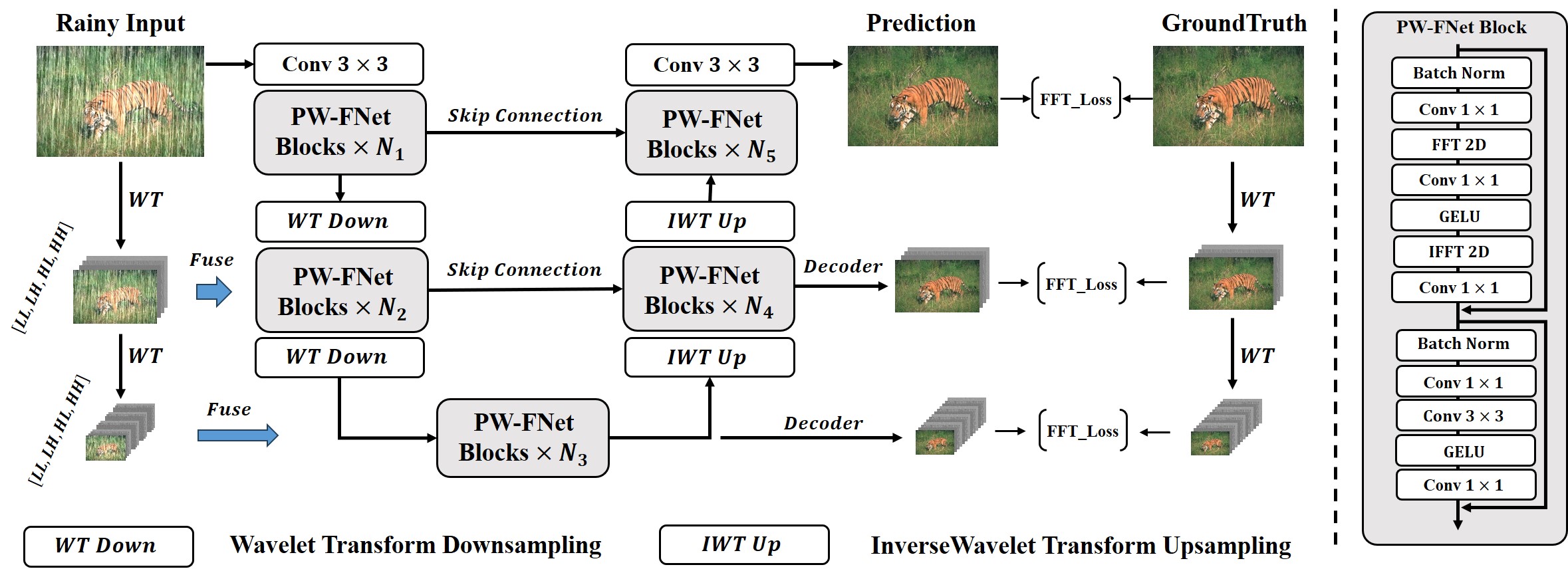}
\caption{Overall framework of our PW-FNet, which integrates wavelet and Fourier transforms into inter-block and intra-block.
}
\label{fig:method}
\vspace{-2mm}
\end{figure*}

\section{Methods}

\subsection{Overall Framework}
Building upon the design principles of NAFNet\cite{NAFNet}, which divides the network architecture into two primary levels: inter-block and intra-block, we propose the Pyramid Wavelet-Fourier Network (PW-FNet) framework, as shown in Fig.\ref{fig:method}, integrating Wavelet-Fourier into the inter-block and intra-block architectural designs. Specifically, at the inter-block level, we introduce a pyramid wavelet multi-input multi-output (PW-MIMO) structure to achieve multi-scale and multi-frequency feature decomposition, while also enabling dynamic adaptive outputs. At the intra-block level, we replace self-attention mechanisms with Fourier transforms, facilitating efficient global modeling. In practice, PW-FNet utilizes a 3-level hierarchical encoder-decoder structure. Given a degraded image \(I\), the network first applies multiple wavelet transformations \(\mathcal{W}\) to generate multi-scale image inputs \(\{I^i, i=1,2,4\}\). Then, 3x3 convolutions are applied to each multi-input to increase the feature dimension and extract corresponding low-level features \(\{f^i, i=1,2,4\}\). These shallow features \(\{f^i, i=1,2,4\}\) are then passed through the 3-level hierarchical encoder-decoder structure. At each level, the spatial resolution is progressively reduced through trainable wavelet transform downsampling in the encoder and increased through trainable inverse wavelet transform upsampling in the decoder. Within each level, the features \(\{f^i, i=1,2,4\}\) are passed through \(N\) PW-FNet blocks, which extract multi-scale latent features \(\{f^i_a, i=1,2,4\}\). Finally, 3x3 convolutions are applied to map \(\{f^i_a, i=1,2,4\}\) to the residual image \(\{r^i, i=1,2,4\}\), and the clear output is obtained as \(\{o^i, i=1,2,4\} = \{I^i + r^i, i=1,2,4\}\). In the following sections, we will provide a detailed explanation of the structural design of PW-FNet at both the inter-block and intra-block levels.


\subsection{Pyramid Wavelet Multi-input Multi-output}
Initially, most restoration baselines \cite{Uformer, Restormer, NAFNet} adopted a single-input single-output (SISO) architecture, where the network processes a single-resolution input and produces a single-resolution output. Recently, multi-input multi-output (MIMO) architectures \cite{sfnet, focalnet} have gained popularity, as they enable multi-scale feature extraction through multiple input and output branches. Compared to SISO, MIMO can be seen as a form of multi-task learning, which improves performance by capturing richer hierarchical representations.

In this work, we propose Pyramid Wavelet Multi-Input Multi-Output (PW-MIMO) architecture by integrating pyramid wavelet transforms into the MIMO framework. Compared to the standard MIMO structure, our design introduces two key enhancements: 1) For multi-input, we incorporate wavelet transforms to generate multiple inputs, which preserves lossless information while producing different pyramid-scale multi-frequency components to separate degradation from background features. 2) For multi-output, we apply inverse wavelet transforms to produce multiple outputs, enabling the network to generate three stages of restored images at different degradation levels. This design further facilitates dynamic adaptive forward inference, allowing us to obtain small, medium, and large model variants with varying parameter and computational budgets from a single trained network, eliminating the need for separate training processes.

\subsection{PW-FNet Block}
As shown in Fig.\ref{fig:method}, the core design of the PW-FNet block lies in achieving efficient global modeling by replacing the self-attention mechanism with Fourier transforms. For the Token Mixer part, given an input feature \( f^1 \), PW-FNet first applies a pointwise convolution to double the channel dimension, resulting in \( f^1_2 \). Then, the upsampled feature \( f^1_2 \) undergoes a global 2D Fourier transform \( \mathcal{F}(f^1_2) \), mapping it to the frequency domain. In the frequency domain, the feature sequentially passes through a pointwise convolution and GELU activation function, yielding the feature \( f^1_3 \). Subsequently, a global 2D inverse Fourier transform \( \mathcal{F}^{-1}(f^1_3) \) maps the features back to the spatial domain, resulting in \( f^1_4 \). Finally, another pointwise convolution is applied to reduce the dimension of the transformed features, producing the output \( f^1_5 \). The token mixer part can be formalized as:

\begin{equation}
f^1_5 = \mathcal{F}^{-1}(\text{GELU}(\mathcal{F}(f^1 * W_{1\times1}) * W_{1\times1})  * W_{1\times1}
\end{equation}

For the feed-forward network (FFN) part, we follow the design of Uformer, where an additional depthwise convolution layer is added to the standard FFN to enhance its local representation capabilities. For an input feature \( f^2 \), PW-FNet begins by applying a pointwise convolution to double the channel dimension, resulting in \( f^2_2 \). This upsampled feature \( f^2_2 \), then passes through a depthwise convolution followed by a GELU activation function. Finally, another pointwise convolution is applied to reduce the feature dimension, producing the final output \( f^2_3 \). The feed-forward network part can be formalized as:

\begin{equation}
f^2_3 = \text{GELU}(f^2 * W_{1\times1} * W_{3\times3}) * W_{1\times1}
\end{equation}


\subsection{Loss Function}
Unlike the commonly used spatial-domain loss, we introduce the Fourier-domain loss to align with our model design, as expressed in Eq.\ref{eq:loss}. For the outputs from different branches, we assign equal weights by summing them.


\begin{align}\label{eq:loss}
    L &=  \sum_{i}\left \| \mathcal{F}(o^i) - \mathcal{F}(g^i)  \right \|_{1}, i=1,2,4
\end{align}

The L1 loss is used to regularize the multi-output set \(\{o^i, i=1,2,4\}\), encouraging it to match the ground truth \(\{g^i, i=1,2,4\}\).

\begin{table*}[b]
\centering
\caption{Quantitative evaluations on synthetic and real datasets. \textbf{Bold} and \underline{underline} indicate the best and second-best results.}\label{tab:1}
\begin{tabular}{cllll|ccl|clcl|clcl|clcl|clcl|clcl}
\hline
\multicolumn{5}{c|}{\multirow{3}{*}{Method}}      & \multicolumn{15}{c|}{Synthetic}  & \multicolumn{4}{c|}{Real}  & \multicolumn{4}{c}{\multirow{2}{*}{Overhead}} \\ \cline{6-24}
\multicolumn{5}{c|}{}        & \multicolumn{3}{c|}{Rain200L\cite{RAIN200}}                                & \multicolumn{4}{c|}{Rain200H\cite{RAIN200}}                                & \multicolumn{4}{c|}{DDN-Data\cite{ddn}}           & \multicolumn{4}{c|}{DID-Data\cite{DID}}      & \multicolumn{4}{c|}{SPA-Data\cite{SPA-Data}}    & \multicolumn{4}{c}{}           \\ \cline{6-28}
\multicolumn{5}{c|}{}              & \multicolumn{1}{c}{PSNR}  & \multicolumn{2}{c|}{SSIM}   & \multicolumn{2}{c}{PSNR}  & \multicolumn{2}{c|}{SSIM}   & \multicolumn{2}{c}{PSNR}  & \multicolumn{2}{c|}{SSIM}   & \multicolumn{2}{c}{PSNR}  & \multicolumn{2}{c|}{SSIM} & \multicolumn{2}{c}{PSNR}  & \multicolumn{2}{c|}{SSIM}  & \multicolumn{2}{c}{\#Param} & \multicolumn{2}{c}{FLOPs}   \\ \hline
\multicolumn{5}{c|}{RESCAN\cite{li2018recurrentRESCAN}}            & \multicolumn{1}{c}{36.09} & \multicolumn{2}{c|}{0.9697} & \multicolumn{2}{c}{26.75} & \multicolumn{2}{c|}{0.8353} & \multicolumn{2}{c}{31.94} & \multicolumn{2}{c|}{0.9345} & \multicolumn{2}{c}{33.38} & \multicolumn{2}{c|}{0.9417} & \multicolumn{2}{c}{38.11} & \multicolumn{2}{c|}{0.9707} & \multicolumn{2}{c}{0.150M}      & \multicolumn{2}{c}{32.12G}      \\
\multicolumn{5}{c|}{PReNet\cite{ren2019progressivePReNet}}   & \multicolumn{1}{c}{37.80} & \multicolumn{2}{c|}{0.9814}     & \multicolumn{2}{c}{29.04} & \multicolumn{2}{c|}{0.8991} & \multicolumn{2}{c}{32.60} & \multicolumn{2}{c|}{0.9459} & \multicolumn{2}{c}{33.17}     & \multicolumn{2}{c|}{0.9481}      & \multicolumn{2}{c}{40.16}     & \multicolumn{2}{c|}{0.9816}      & \multicolumn{2}{c}{0.169M} & \multicolumn{2}{c}{66.25G} \\
\multicolumn{5}{c|}{MSPFN\cite{jiang2020multiMSPFN}}    & \multicolumn{1}{c}{38.58} & \multicolumn{2}{c|}{0.9827}      & \multicolumn{2}{c}{29.36} & \multicolumn{2}{c|}{0.9034} & \multicolumn{2}{c}{32.99} & \multicolumn{2}{c|}{0.9333} & \multicolumn{2}{c}{33.72} & \multicolumn{2}{c|}{0.9550} & \multicolumn{2}{c}{43.43} & \multicolumn{2}{c|}{0.9843} & \multicolumn{2}{c}{20.89M} & \multicolumn{2}{c}{595.5G} \\
\multicolumn{5}{c|}{RCDNet\cite{wang2020modelRCDNet}} & \multicolumn{1}{c}{39.17} & \multicolumn{2}{c|}{0.9885}  & \multicolumn{2}{c}{30.24} & \multicolumn{2}{c|}{0.9048} & \multicolumn{2}{c}{33.04} & \multicolumn{2}{c|}{0.9472} & \multicolumn{2}{c}{34.08} & \multicolumn{2}{c|}{0.9532} & \multicolumn{2}{c}{43.36} & \multicolumn{2}{c|}{0.9831} & \multicolumn{2}{c}{2.958M} & \multicolumn{2}{c}{194.5G} \\
\multicolumn{5}{c|}{MPRNet\cite{zamir2021multiMPRNet}}    & \multicolumn{1}{c}{39.47} & \multicolumn{2}{c|}{0.9825}       & \multicolumn{2}{c}{30.67} & \multicolumn{2}{c|}{0.9110} & \multicolumn{2}{c}{33.10} & \multicolumn{2}{c|}{0.9347} & \multicolumn{2}{c}{33.99} & \multicolumn{2}{c|}{0.9590} & \multicolumn{2}{c}{43.64} & \multicolumn{2}{c|}{0.9844} & \multicolumn{2}{c}{3.637M} & \multicolumn{2}{c}{548.7G} \\
\multicolumn{5}{c|}{DualGCN\cite{fu2021rainDualGCN}}    & \multicolumn{1}{c}{40.73} & \multicolumn{2}{c|}{0.9886}     & \multicolumn{2}{c}{31.15} & \multicolumn{2}{c|}{0.9125} & \multicolumn{2}{c}{33.01} & \multicolumn{2}{c|}{0.9489} & \multicolumn{2}{c}{34.37} & \multicolumn{2}{c|}{0.9620} & \multicolumn{2}{c}{44.18} & \multicolumn{2}{c|}{0.9902} & \multicolumn{2}{c}{2.73M} & \multicolumn{2}{c}{-} \\
\multicolumn{5}{c|}{SPDNet\cite{yi2021structureSPDNet}}   & \multicolumn{1}{c}{40.50} & \multicolumn{2}{c|}{0.9875}     & \multicolumn{2}{c}{31.28} & \multicolumn{2}{c|}{0.9207} & \multicolumn{2}{c}{33.15}     & \multicolumn{2}{c|}{0.9457}      & \multicolumn{2}{c}{34.57}      & \multicolumn{2}{c|}{0.9560}       & \multicolumn{2}{c}{43.20}      & \multicolumn{2}{c|}{0.9871}       & \multicolumn{2}{c}{2.982M} & \multicolumn{2}{c}{96.29G} \\
\multicolumn{5}{c|}{Uformer\cite{wang2022uformerUformer}}    & \multicolumn{1}{c}{40.20} & \multicolumn{2}{c|}{0.9860}     & \multicolumn{2}{c}{30.80} & \multicolumn{2}{c|}{0.9105} & \multicolumn{2}{c}{33.95} & \multicolumn{2}{c|}{0.9545} & \multicolumn{2}{c}{35.02}      & \multicolumn{2}{c|}{0.9621}       & \multicolumn{2}{c}{46.13}      & \multicolumn{2}{c|}{0.9913}       & \multicolumn{2}{c}{20.60M}       & \multicolumn{2}{c}{41.09G}       \\
\multicolumn{5}{c|}{Restormer\cite{zamir2022restormerRestormer}}   & \multicolumn{1}{c}{40.99} & \multicolumn{2}{c|}{0.9890}    & \multicolumn{2}{c}{32.00} & \multicolumn{2}{c|}{0.9329} & \multicolumn{2}{c}{34.20} & \multicolumn{2}{c|}{0.9571}  & \multicolumn{2}{c}{35.29}     & \multicolumn{2}{c|}{0.9641}      & \multicolumn{2}{c}{47.98}     & \multicolumn{2}{c|}{0.9921}      & \multicolumn{2}{c}{26.10M} & \multicolumn{2}{c}{141.0G} \\ 
\multicolumn{5}{c|}{IDT\cite{xiao2022imageIDT}}    & \multicolumn{1}{c}{40.74} & \multicolumn{2}{c|}{0.9884}    & \multicolumn{2}{c}{32.10}      & \multicolumn{2}{c|}{0.9344}       & \multicolumn{2}{c}{33.84}      & \multicolumn{2}{c|}{0.9549}       & \multicolumn{2}{c}{34.89}      & \multicolumn{2}{c|}{0.9623}       & \multicolumn{2}{c}{47.35}      & \multicolumn{2}{c|}{0.9930}       & \multicolumn{2}{c}{16.39M}       & \multicolumn{2}{c}{58.44G}       \\
\multicolumn{5}{c|}{HCT-FFN\cite{chen2023hybridHCT-FFN}}   & \multicolumn{1}{c}{39.70} & \multicolumn{2}{c|}{0.9850}     & \multicolumn{2}{c}{31.51}      & \multicolumn{2}{c|}{0.9100}       & \multicolumn{2}{c}{33.00}      & \multicolumn{2}{c|}{0.9502}       & \multicolumn{2}{c}{33.96}      & \multicolumn{2}{c|}{0.9592}       & \multicolumn{2}{c}{45.79}      & \multicolumn{2}{c|}{0.9898}       & \multicolumn{2}{c}{0.874M}       & \multicolumn{2}{c}{80.25G}       \\
\multicolumn{5}{c|}{DRSformer\cite{chen2023learningDRSformer}}    & \multicolumn{1}{c}{41.23} & \multicolumn{2}{c|}{0.9894}    & \multicolumn{2}{c}{32.17}      & \multicolumn{2}{c|}{0.9326}       & \multicolumn{2}{c}{34.35}      & \multicolumn{2}{c|}{0.9588}       & \multicolumn{2}{c}{35.35}      & \multicolumn{2}{c|}{0.9646}       & \multicolumn{2}{c}{48.54}      & \multicolumn{2}{c|}{0.9924}       & \multicolumn{2}{c}{33.70M}       & \multicolumn{2}{c}{242.9G}       \\
\multicolumn{5}{c|}{NeRD-Rain\cite{chen2024NeRD}}    & \multicolumn{1}{c}{41.71} & \multicolumn{2}{c|}{0.9903}    & \multicolumn{2}{c}{32.40}      & \multicolumn{2}{c|}{\underline{0.9373}}       & \multicolumn{2}{c}{\underline{34.45}}      & \multicolumn{2}{c|}{0.9596}       & \multicolumn{2}{c}{\underline{35.53}}      & \multicolumn{2}{c|}{\underline{0.9659}}       & \multicolumn{2}{c}{\underline{49.58}}      & \multicolumn{2}{c|}{\textbf{0.9940}}       & \multicolumn{2}{c}{22.89M}       & \multicolumn{2}{c}{156.3G}       \\
\multicolumn{5}{c|}{FADformer\cite{gao2024FADformer}}  & \multicolumn{1}{c}{41.80} & \multicolumn{2}{c|}{\underline{0.9906}} & \multicolumn{2}{c}{\underline{32.48}}      & \multicolumn{2}{c|}{0.9359}      & \multicolumn{2}{c}{34.42}      & \multicolumn{2}{c|}{\underline{0.9602}}       & \multicolumn{2}{c}{35.48}      & \multicolumn{2}{c|}{0.9657}       & \multicolumn{2}{c}{49.21}      & \multicolumn{2}{c|}{0.9934}       & \multicolumn{2}{c}{6.958M}       & \multicolumn{2}{c}{48.51G}       \\ \hline
\multicolumn{5}{c|}{(Ours)PW-FNet-S}    & \multicolumn{1}{c}{41.64} & \multicolumn{2}{c|}{0.9900}    & \multicolumn{2}{c}{31.97}      & \multicolumn{2}{c|}{0.9305}       & \multicolumn{2}{c}{34.13}      & \multicolumn{2}{c|}{0.9571}       & \multicolumn{2}{c}{35.20}      & \multicolumn{2}{c|}{0.9640}       & \multicolumn{2}{c}{48.38}      & \multicolumn{2}{c|}{0.9934}       & \multicolumn{2}{c}{0.719M}       & \multicolumn{2}{c}{16.64G}       \\
\multicolumn{5}{c|}{(Ours)PW-FNet-M}    & \multicolumn{1}{c}{\underline{41.82}} & \multicolumn{2}{c|}{0.9904}    & \multicolumn{2}{c}{32.37}      & \multicolumn{2}{c|}{0.9342}       & \multicolumn{2}{c}{34.31}      & \multicolumn{2}{c|}{0.9591}       & \multicolumn{2}{c}{35.39}      & \multicolumn{2}{c|}{0.9652}       & \multicolumn{2}{c}{49.02}      & \multicolumn{2}{c|}{0.9936}       & \multicolumn{2}{c}{1.196M}       & \multicolumn{2}{c}{22.38G}       \\
\multicolumn{5}{c|}{\textbf{(Ours)PW-FNet-L}}  & \multicolumn{1}{c}{\textbf{42.23}} & \multicolumn{2}{c|}{\textbf{0.9915}} & \multicolumn{2}{c}{\textbf{32.88}}      & \multicolumn{2}{c|}{\textbf{0.9413}}       & \multicolumn{2}{c}{\textbf{34.48}}      & \multicolumn{2}{c|}{\textbf{0.9606}}       & \multicolumn{2}{c}{\textbf{35.54}}      & \multicolumn{2}{c|}{\textbf{0.9663}}       & \multicolumn{2}{c}{\textbf{49.79}}      & \multicolumn{2}{c|}{\underline{0.9936}}       & \multicolumn{2}{c}{1.442M}       & \multicolumn{2}{c}{33.56G}       \\ \hline
\end{tabular}
\end{table*}

\section{Experiments}
\subsection{Experimental settings}
\noindent \textbf{Restoration Datasets.} We evaluate the effectiveness of our method across seven different image restoration tasks. For the image deraining task, we utilize classical low-resolution rain-streak removal datasets, including Rain200L/H \cite{yang2017deepRain200}, DID-Data \cite{zhang2018densityDID}, and DDN-Data \cite{fu2017removingDDN}, as well as additional datasets for broader generalizability. These include a raindrop removal dataset (Raindrop \cite{RaindropAttn}), a real-world deraining dataset (SPA-Data \cite{wang2019spatialSPA}) and a high-resolution deraining dataset (4K-Rain13K \cite{chen2024UDRMixer}). For image super-resolution, we use the widely recognized DIV2K\cite{DIV2K} dataset for training, and the benchmarks Set5\cite{set5}, Set14\cite{set14}, BSD100\cite{bsd100}, Manga109\cite{manga109} and Urban100\cite{urban100} for testing. For motion deblurring, we employ the GoPro\cite{GOPRO} dataset. For image dehazing, we use NH-HAZE \cite{nhHAZE} and DENSE-HAZE \cite{denseHAZE}; for image desnowing, we adopt SRRS \cite{JSTASRSRRS} and Snow100K \cite{desnownesnow100k}; for underwater enhancement, we utilize LSUI \cite{U-shapeTranslsui}; and for low-light enhancement, we employ LOL-v2 \cite{sparselolv2}. For detailed descriptions of each dataset, please refer to the supplementary material.

\noindent \textbf{Implementation Details.} For the configuration of PW-FNet, we utilize the outputs of the last three stages as nodes to categorize PW-FNet into small, medium and large scales. PW-FNet is trained from scratch on each dataset using PyTorch \cite{paszke2019pytorchPytorch} on four NVIDIA GeForce RTX 3090 GPUs. We employ the AdamW optimizer \cite{loshchilov2018fixingAdamW} with hyperparameters $\beta_1 = 0.9$ and $\beta_2 = 0.999$. Training is conducted with a batch size of 24 and a patch size of $256 \times 256$ for 500K iterations. The learning rate is initially set to $1 \times 10^{-3}$ and gradually decayed to $1 \times 10^{-6}$ following a cosine annealing schedule \cite{loshchilov2016sgdrCosine}. Training patches are randomly cropped and further augmented through rotation and flipping.

\begin{figure*}[t]
\setlength{\abovecaptionskip}{0cm}
\setlength{\belowcaptionskip}{-0.cm}
	\centering
	\includegraphics[width=\linewidth]{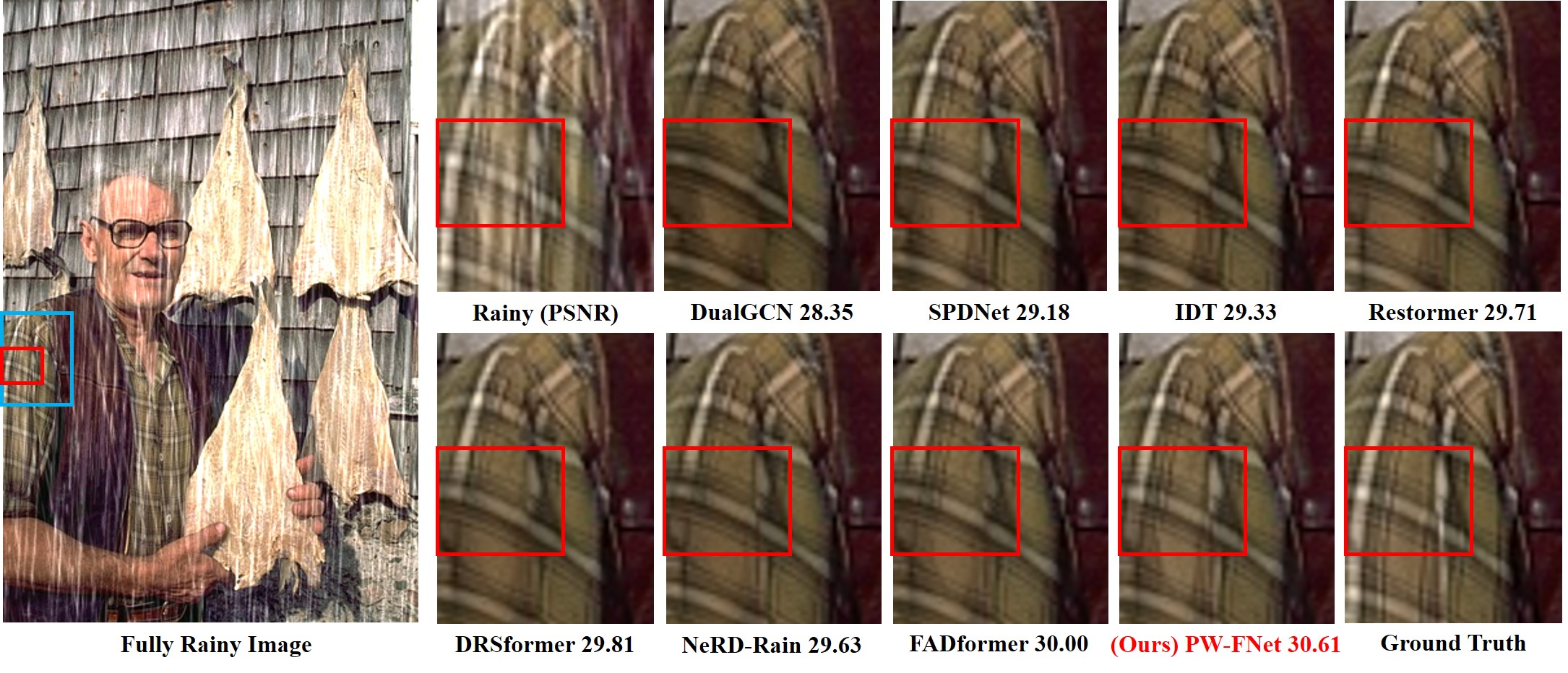}
	\caption{Quantitative evaluation results on the Rain200H\cite{yang2017deepRain200} dataset.
	}
	\label{fig:rain200h}
    \vspace{-3mm}
\end{figure*}

\begin{figure*}[t]
\setlength{\abovecaptionskip}{0cm}
\setlength{\belowcaptionskip}{-0.cm}
	\centering
	\includegraphics[width=\linewidth]{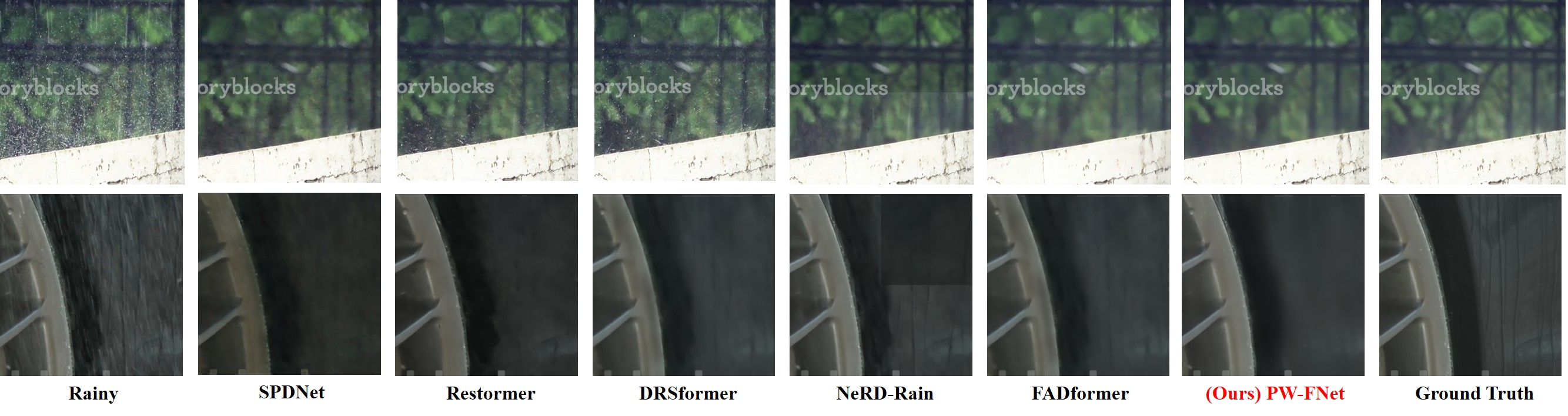}
	\caption{Quantitative evaluation results on the real-world SPA-Data\cite{wang2019spatialSPA} dataset.
	}
	\label{fig:spa}
    \vspace{-1mm}
\end{figure*}

\begin{figure*}[t]
\setlength{\abovecaptionskip}{0cm}
\setlength{\belowcaptionskip}{-0.cm}
	\centering
	\includegraphics[width=\linewidth]{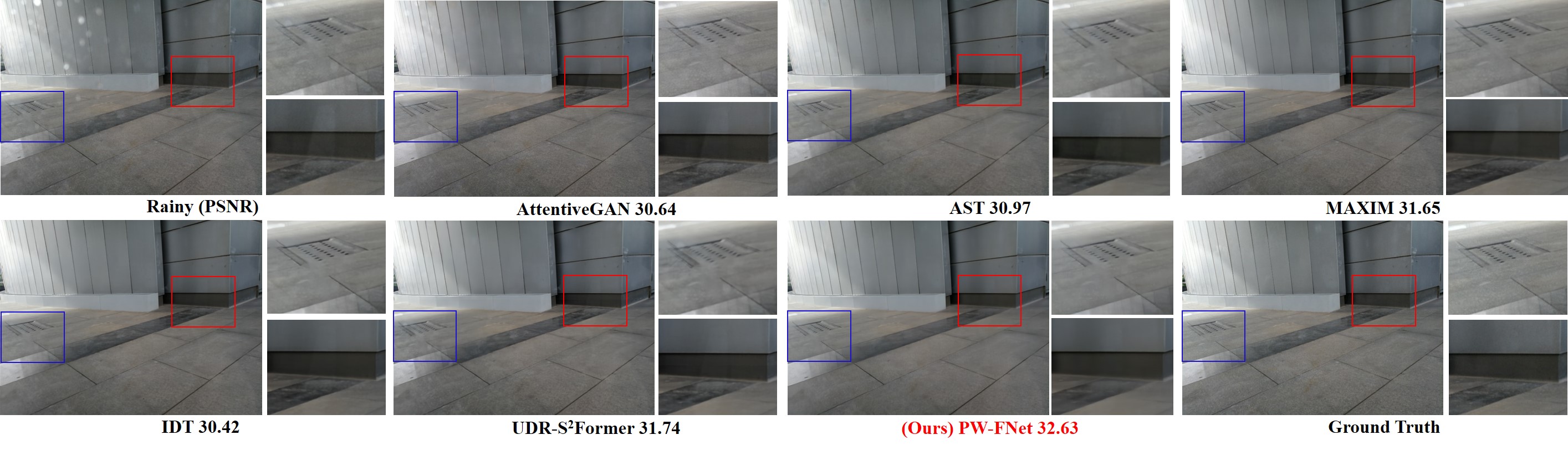}
	\caption{Quantitative evaluation results on the Raindrop\cite{AttentiveGANraindrop} dataset.
	}
	\label{fig:raindrop}
    \vspace{-3mm}
\end{figure*}

\begin{figure*}[t]
\setlength{\abovecaptionskip}{0cm}
\setlength{\belowcaptionskip}{-0.cm}
	\centering
	\includegraphics[width=\linewidth]{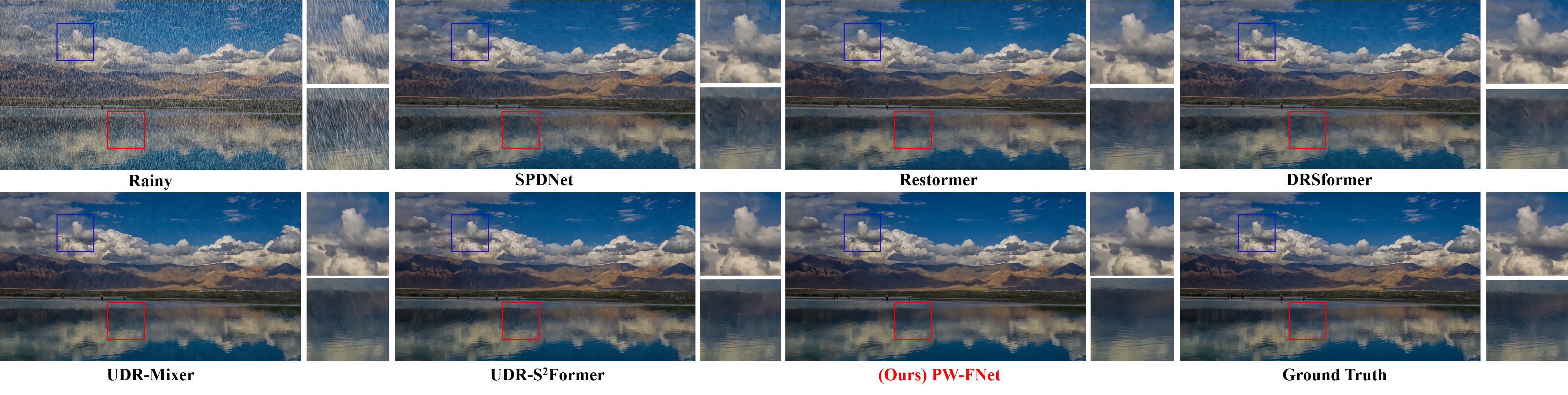}
	\caption{Quantitative evaluation results on the high-resolution 4K-Rain13K\cite{chen2024UDRMixer} dataset.
	}
	\label{fig:4k}
\end{figure*}

\subsection{Experimental results on image deraining}
\noindent \textbf{Synthetic datasets results.} Tab.\ref{tab:1} presents the quantitative results on four synthetic deraining datasets: Rain200H\cite{yang2017deepRain200}, Rain200L\cite{yang2017deepRain200}, DDN-Data\cite{fu2017removingDDN}, and DID-Data\cite{zhang2018densityDID}. Compared to state-of-the-art methods, our PW-FNet-L achieves the best performance across all four datasets in terms of PSNR and SSIM. Notably, compared to NeRD-Rain\cite{chen2024NeRD}, our model reduces the parameter size to only 1/16th and the computational cost to 1/5th, yet still outperforms it by 0.5dB in PSNR on Rain200H and Rain200L. Meanwhile, our medium (PW-FNet-M) and small (PW-FNet-S) variants remain highly competitive, delivering strong performance with significantly fewer parameters and lower computational cost. Fig.\ref{fig:rain200h} presents the qualitative deraining results on the Rain200H dataset. Our model preserves finer texture details, while other models tend to over-remove rain, mistakenly treating white clothing straps as rain and eliminating them.

\noindent \textbf{Real-world datasets results.} The last few columns of Tab.\ref{tab:1} present the qualitative results on the real-world deraining dataset SPA-Data\cite{wang2019spatialSPA}. Compared to other methods, our PW-FNet achieves the best PSNR while maintaining extremely low model complexity. Fig.\ref{fig:spa} shows the qualitative deraining results on the SPA-Data dataset. It is evident that our model delivers the best visual quality in real-world deraining tasks, effectively removing most of the rain while preserving more details. In contrast, Restormer\cite{zamir2022restormerRestormer} and DRSformer\cite{chen2023learningDRSformer} fail to fully remove the rain, while NeRD-Rain\cite{chen2024NeRD} exhibits significant artifact problems.

\begin{table}[h]
\centering
\caption{Quantitative evaluations on raindrop removal.}\label{tab:2}
\resizebox{0.48\textwidth}{!}{
\begin{tabular}{clllllll|clcl|clcl}
\hline
\multicolumn{8}{c|}{\multirow{2}{*}{Methods}} & \multicolumn{4}{c|}{Raindrop-A\cite{AttentiveGANraindrop}}                        & \multicolumn{4}{c}{Raindrop-B\cite{AttentiveGANraindrop}}                                                \\ \cline{9-16} 
\multicolumn{8}{c|}{}                         & \multicolumn{2}{c}{PSNR}  & \multicolumn{2}{c|}{SSIM}  & \multicolumn{2}{c}{PSNR}  & \multicolumn{2}{c}{SSIM} \\ \hline
\multicolumn{8}{c|}{pix2pix\cite{pix2pix}}                  & \multicolumn{2}{c}{28.02} & \multicolumn{2}{c|}{0.855} & \multicolumn{2}{c}{-}     & \multicolumn{2}{c}{-}        \\
\multicolumn{8}{c|}{DuRN\cite{DuRN}}                     & \multicolumn{2}{c}{31.24} & \multicolumn{2}{c|}{0.926} & \multicolumn{2}{c}{25.32} & \multicolumn{2}{c}{0.817}     \\
\multicolumn{8}{c|}{RaindropAttn\cite{RaindropAttn}}             & \multicolumn{2}{c}{31.44} & \multicolumn{2}{c|}{0.926} & \multicolumn{2}{c}{-}     & \multicolumn{2}{c}{-}          \\
\multicolumn{8}{c|}{AttentiveGAN\cite{AttentiveGANraindrop}}             & \multicolumn{2}{c}{31.59} & \multicolumn{2}{c|}{0.917} & \multicolumn{2}{c}{25.05} & \multicolumn{2}{c}{0.811}       \\
\multicolumn{8}{c|}{IDT\cite{xiao2022imageIDT}}                      & \multicolumn{2}{c}{31.87} & \multicolumn{2}{c|}{0.931} & \multicolumn{2}{c}{-}     & \multicolumn{2}{c}{-}         \\
\multicolumn{8}{c|}{MAXIM\cite{MAXIM}}                 & \multicolumn{2}{c}{31.87} & \multicolumn{2}{c|}{0.935} & \multicolumn{2}{c}{25.74} & \multicolumn{2}{c}{0.827}     \\
\multicolumn{8}{c|}{RainDropDiff128\cite{RainDropDiff128}}          & \multicolumn{2}{c}{32.43} & \multicolumn{2}{c|}{0.933} & \multicolumn{2}{c}{-}     & \multicolumn{2}{c}{-}         \\ 
\multicolumn{8}{c|}{UDR-$\mathrm{S^2}$Former\cite{chen2023UDRS2Former}}                         & \multicolumn{2}{c}{32.64}      & \multicolumn{2}{c|}{0.943}      & \multicolumn{2}{c}{26.92}      & \multicolumn{2}{c}{0.8317}          \\ 
\multicolumn{8}{c|}{AST\cite{zhou2024AST}}                         & \multicolumn{2}{c}{32.45}      & \multicolumn{2}{c|}{0.937}      & \multicolumn{2}{c}{24.99}      & \multicolumn{2}{c}{0.8058}          \\ \hline
\multicolumn{8}{c|}{(Ours)PW-FNet-S}                         & \multicolumn{2}{c}{\textbf{33.27}}      & \multicolumn{2}{c|}{0.9445}      & \multicolumn{2}{c}{\textbf{27.10}}      & \multicolumn{2}{c}{0.8332}          \\ 
\multicolumn{8}{c|}{(Ours)PW-FNet-M}                         & \multicolumn{2}{c}{33.04}      & \multicolumn{2}{c|}{\underline{0.9457}}      & \multicolumn{2}{c}{27.02}      & \multicolumn{2}{c}{\underline{0.8342}}          \\ 
\multicolumn{8}{c|}{(Ours)PW-FNet-L}                         & \multicolumn{2}{c}{\underline{33.21}}      & \multicolumn{2}{c|}{\textbf{0.9469}}      & \multicolumn{2}{c}{\underline{27.04}}      & \multicolumn{2}{c}{\textbf{0.8343}}          \\ \hline
\end{tabular}}
\end{table}

\noindent \textbf{Raindrop datasets results.} Tab.\ref{tab:2} presents the qualitative results on the Raindrop\cite{AttentiveGANraindrop} dataset. Compared to state-of-the-art methods, our PW-FNet model achieves the best performance in terms of PSNR and SSIM across both test datasets. Interestingly, unlike the rain streak task, the small version of our model achieves the best PSNR on the raindrop dataset, outperforming the large version. We attribute this to the different degradation mechanisms in the rain streak and raindrop tasks. Raindrop degradation has a more pronounced impact on high-frequency information, resembling a blur effect. In such cases, smaller-scale outputs tend to perform better. Fig.\ref{fig:raindrop} presents the qualitative deraining results on the Raindrop dataset. It can be seen that our model restores images with smoother structures and better consistency, demonstrating its effectiveness in handling raindrop-based degradation.

\begin{table}[h]
\renewcommand{\arraystretch}{1}
\centering
\caption{Quantitative evaluations on the High-Resolution deraining task. 
Latency is measured on 1024×1024 resolution.}\label{tab:3}
\resizebox{0.45\textwidth}{!}{
\begin{tabular}{clllllll|clclcl}
\hline
\multicolumn{8}{c|}{\multirow{2}{*}{Methods}} & \multicolumn{6}{c}{4K-Rain13k\cite{chen2024UDRMixer}}                 \\ \cline{9-14} 
\multicolumn{8}{c|}{}                         & \multicolumn{2}{c}{PSNR}  & \multicolumn{2}{c|}{SSIM}  & \multicolumn{2}{c}{Latency} \\ \hline
\multicolumn{8}{c|}{DSC\cite{luo2015removingDSC}}  & \multicolumn{2}{c}{22.93} & \multicolumn{2}{c|}{0.6299} & \multicolumn{2}{c}{-}    \\
\multicolumn{8}{c|}{LPNet\cite{fu2019LPNet}}  & \multicolumn{2}{c}{27.86} & \multicolumn{2}{c|}{0.8924} & \multicolumn{2}{c}{-}    \\
\multicolumn{8}{c|}{JORDER-E\cite{yang2019Jorder}}  & \multicolumn{2}{c}{30.46} & \multicolumn{2}{c|}{0.9117} & \multicolumn{2}{c}{-}    \\
\multicolumn{8}{c|}{RCDNet\cite{wang2020modelRCDNet}}  & \multicolumn{2}{c}{30.83} & \multicolumn{2}{c|}{0.9212} & \multicolumn{2}{c}{-}  \\
\multicolumn{8}{c|}{SPDNet\cite{yi2021structureSPDNet}}  & \multicolumn{2}{c}{31.81} & \multicolumn{2}{c|}{0.9223} & \multicolumn{2}{c}{391ms}    \\
\multicolumn{8}{c|}{IDT\cite{xiao2022imageIDT}}  & \multicolumn{2}{c}{32.91} & \multicolumn{2}{c|}{0.9479} & \multicolumn{2}{c}{1935ms}    \\
\multicolumn{8}{c|}{Restormer\cite{zamir2022restormerRestormer}}  & \multicolumn{2}{c}{33.02} & \multicolumn{2}{c|}{0.9335} & \multicolumn{2}{c}{1145ms}    \\
\multicolumn{8}{c|}{DRSformer\cite{chen2023learningDRSformer}}  & \multicolumn{2}{c}{32.96} & \multicolumn{2}{c|}{0.9334} & \multicolumn{2}{c}{2682ms}    \\
\multicolumn{8}{c|}{UDR-$\mathrm{S^2}$Former\cite{chen2023UDRS2Former}}  & \multicolumn{2}{c}{33.36} & \multicolumn{2}{c|}{0.9458} & \multicolumn{2}{c}{547ms}    \\
\multicolumn{8}{c|}{UDR-Mixer\cite{chen2024UDRMixer}}  & \multicolumn{2}{c}{34.30} & \multicolumn{2}{c|}{0.9505} & \multicolumn{2}{c}{77ms}    \\ \hline
\multicolumn{8}{c|}{(Ours)PW-FNet-S}  & \multicolumn{2}{c}{35.02} & \multicolumn{2}{c|}{0.9565} & \multicolumn{2}{c}{39ms}    \\
\multicolumn{8}{c|}{(Ours)PW-FNet-M}  & \multicolumn{2}{c}{\underline{35.77}} & \multicolumn{2}{c|}{\underline{0.9588}} & \multicolumn{2}{c}{50ms}    \\ 
\multicolumn{8}{c|}{(Ours)PW-FNet-L}  & \multicolumn{2}{c}{\textbf{35.93}} & \multicolumn{2}{c|}{\textbf{0.9607}} & \multicolumn{2}{c}{81ms}    \\ \hline
\end{tabular}}
\vspace{-0.mm}
\end{table}

\noindent \textbf{High-resolution datasets results.} Tab.\ref{tab:3} presents the quantitative results for the high-resolution image deraining task. Compared to other methods, our model demonstrates outstanding performance in high-resolution deraining, with even the small version (PW-FNet-S) achieving the best metrics in terms of PSNR and SSIM. Notably, our PW-FNet-L surpasses UDR-Mixer\cite{chen2024UDRMixer} by 1.63dB in PSNR, highlighting its superior restoration capability. Furthermore, we evaluate the inference time of different models at a resolution of 1024×1024, as shown in latency column of Tab.\ref{tab:3}. Our PW-FNet-S achieves an inference time of 39ms(25 FPS), whereas Restormer\cite{zamir2022restormerRestormer} requires 1145ms(only 0.87 FPS), demonstrating the significant efficiency of our model. Fig.\ref{fig:4k} presents the qualitative results on the high-resolution deraining task. It is evident that our method achieves the best visual quality across the entire high-resolution image, removing rain more effectively. In contrast, other methods fail to completely eliminate rain streaks in certain regions, leaving visible residual rain patterns.

\begin{table}[h]
\renewcommand{\arraystretch}{1}
\centering
\caption{Quantitative evaluations on the motion deblurring task. Params and FLOPs are measured on 128×128 resolution.}\label{tab:blur}
\resizebox{0.48\textwidth}{!}{
\begin{tabular}{c|c|c|cc}
\hline
\multirow{2}{*}{Methods}             & \multirow{2}{*}{\#Param} & \multirow{2}{*}{FLOPs} & \multicolumn{2}{c}{GoPro\cite{GOPRO}}                             \\ \cline{4-5} 
                                     &                        &                        & PSNR                      & SSIM                      \\ \hline
MIMO+\cite{mimo}                                & 16.10M                 & 38.64G                 & 32.45                     & 0.957                     \\
MPRNet\cite{MPRNet}                               & 20.13M                 & 194.42G                & 32.66                     & 0.959                     \\
Restormer\cite{Restormer}                            & 26.13M                 & 35.31G                 & 32.92                     & 0.961                     \\
Uformer\cite{Uformer}                              & 50.88M                 & 22.36G                 & 33.06                     & \underline{0.967}                     \\
NAFNet\cite{NAFNet}                               & 67.89M                 & 15.85G                 & 33.71                     & 0.967                     \\
SFNet\cite{sfnet}                                & 13.27M                 & 31.38G                 & 33.27                     & 0.963                     \\
X-Restormer\cite{xrestormer}                          & 26.0M                  & 41.08G                 & 33.44                     & 0.946                     \\
ACL\cite{acl}                                 & 4.6M                 & 13.75G                 & 33.25                     & 0.964                     \\
MaIR\cite{mair}                                 & 26.29M                 & 49.29G                 & 33.69                     & 0.966                     \\ \hline
(Ours)PW-FNet-S                      & 5.04M                       &      7.97G                  & 33.30                     & 0.964                     \\
(Ours)PW-FNet-M & 5.78M  & 10.30G  & \underline{33.72} & 0.966 \\
(Ours)PW-FNet-L & 5.98M  &  12.71G & \textbf{34.03} & \textbf{0.968} \\ \hline
\end{tabular}}
\vspace{-3mm}
\end{table}

\begin{table*}[t]
\setlength{\abovecaptionskip}{0.0cm}
\setlength{\belowcaptionskip}{0.0cm}
\renewcommand{\arraystretch}{1}
\centering
\begin{minipage}[c]{\textwidth}
\centering
\captionof{table}{Quantitative evaluations on the super-resolution task, with Params and MACs measured following \cite{mambair}.}\label{tab:super}
\begin{tabular}{c|c|c|c|cl|cl|cl|cl}
\hline
\multirow{2}{*}{Method} & \multirow{2}{*}{Scale} & \multirow{2}{*}{Params} & \multirow{2}{*}{MACs} & \multicolumn{2}{c|}{BSD100\cite{bsd100}}       & \multicolumn{2}{c|}{Urban100\cite{urban100}}     & \multicolumn{2}{c|}{Manga109\cite{manga109}}     & \multicolumn{2}{c}{Average}       \\
                        &                        &                         &                       & \multicolumn{1}{l}{PSNR} & SSIM   & \multicolumn{1}{l}{PSNR} & SSIM   & \multicolumn{1}{l}{PSNR} & SSIM   & \multicolumn{1}{l}{PSNR} & SSIM   \\ \hline
CARN\cite{carn}                    & \multirow{7}{*}{x2}    & 1,592K                  & 222.8G                & 32.09                    & 0.8978 & 31.92                    & 0.9256 & 38.36                    & 0.9765 & 34.27                    & 0.9342 \\
ELAN\cite{elan}                    &                        & 621K                    & 203.1G                & 32.30                    & 0.9012 & 32.76                    & 0.9340 & 39.11                    & 0.9782 & 34.86                    & 0.9385 \\
SwinIR-light\cite{swinir}            &                        & 910K                    & 244.2G                & 32.31                    & 0.9012 & 32.76                    & 0.9340 & 39.12                    & 0.9783 & 34.87                    & 0.9386 \\
SRFormer-light\cite{srformer}          &                        & 853K                    & 236.3G                & \underline{32.36}                    & \underline{0.9019} & \underline{32.91}                    & \underline{0.9353} & \underline{39.28}                    & \textbf{0.9785} & \underline{34.98}                    & \underline{0.9393} \\
MambaIR-light\cite{mambair}           &                        & 905K                    & 334.2G                & 32.31                    & 0.9013 & 32.85                    & 0.9349 & 39.20                    & 0.9782 & 34.92                    & 0.9388 \\
MaIR\cite{mair}                    &                        & 878K                    & 207.8G                & 32.31                    & 0.9013 & 32.89                    & 0.9346 & 39.22                    & 0.9778 & 34.94                    & 0.9386 \\
(Ours)PW-FNet           &                        & 624K                    & 118.9G                & \textbf{32.37}                    & \textbf{0.9024} & \textbf{33.04}                    & \textbf{0.9369} & \textbf{39.36}                    & \underline{0.9783} & \textbf{35.05}                    & \textbf{0.9398} \\ \hline
CARN\cite{carn}                    & \multirow{7}{*}{x3}    & 1,592K                  & 118.8G                & 29.06                    & 0.8034 & 28.06                    & 0.8493 & 33.50                    & 0.9440 & 30.36                    & 0.8676 \\
ELAN\cite{elan}                    &                        & 629K                    & 90.1G                 & 29.21                    & 0.8081 & 28.69                    & 0.8624 & 34.00                    & 0.9478 & 30.78                    & 0.8745 \\
SwinIR-light\cite{swinir}            &                        & 918K                    & 111.2G                & 29.20                    & 0.8082 & 28.66                    & 0.8624 & 33.98                    & 0.9478 & 30.76                    & 0.8746 \\
SRFormer-light\cite{srformer}          &                        & 861K                    & 105.4G                & \underline{29.26}                    & \underline{0.8099} & 28.81                    & \underline{0.8655} & 34.19                    & \underline{0.9489} & 30.90                    & \underline{0.8764} \\
MambaIR-light\cite{mambair}           &                        & 913K                    & 148.5G                & 29.23                    & 0.8084 & 28.70                    & 0.8631 & 34.12                    & 0.9479 & 30.83                    & 0.8749 \\
MaIR\cite{mair}                    &                        & 886K                    & 93.0G                 & 29.25                    & 0.8088 & \underline{28.83}                    & 0.8651 & \underline{34.21}                    & 0.9484 & \underline{30.91}                    & 0.8758 \\
(Ours)PW-FNet           &                        & 648K                    & 56.0G                 & \textbf{29.30}                    & \textbf{0.8108} & \textbf{28.85}                    & \textbf{0.8658} & \textbf{34.39}                    & \textbf{0.9492} & \textbf{30.98}                    & \textbf{0.8768} \\ \hline
CARN\cite{carn}                    & \multirow{7}{*}{x4}    & 1,592K                  & 90.9G                 & 27.58                    & 0.7349 & 26.07                    & 0.7837 & 30.47                    & 0.9084 & 28.19                    & 0.8118 \\
ELAN\cite{elan}                    &                        & 640K                    & 54.1G                 & 27.69                    & 0.7406 & 26.54                    & 0.7982 & 30.92                    & 0.9150 & 28.53                    & 0.8204 \\
SwinIR-light\cite{swinir}            &                        & 930K                    & 63.6G                 & 27.69                    & 0.7406 & 26.47                    & 0.7980 & 30.92                    & 0.9151 & 28.51                    & 0.8204 \\
SRFormer-light\cite{srformer}          &                        & 873K                    & 62.8G                 & \underline{27.73}                    & \underline{0.7422} & \underline{26.67}                    & \textbf{0.8032} & \underline{31.17}                    & \textbf{0.9165} & \underline{28.67}                    & \underline{0.8230} \\
MambaIR-light\cite{mambair}           &                        & 924K                    & 84.6G                 & 27.68                    & 0.7400 & 26.52                    & 0.7983 & 30.94                    & 0.9135 & 28.53                    & 0.8197 \\
MaIR\cite{mair}                    &                        & 897K                    & 53.1G                 & 27.71                    & 0.7414 & 26.60                    & 0.8013 & 31.13                    & 0.9161 & 28.63                    & 0.8220 \\
(Ours)PW-FNet           &                        & 644K                    & 35.2G                 & \textbf{27.77}                    & \textbf{0.7430} & \textbf{26.68}                    & \underline{0.8029} & \textbf{31.22}                    & \textbf{0.9165} & \textbf{28.69}                    &\textbf{0.8231} \\ \hline
\end{tabular}
\vspace{1mm}
\end{minipage}

\begin{minipage}[c]{\textwidth}
\includegraphics[width=\linewidth]{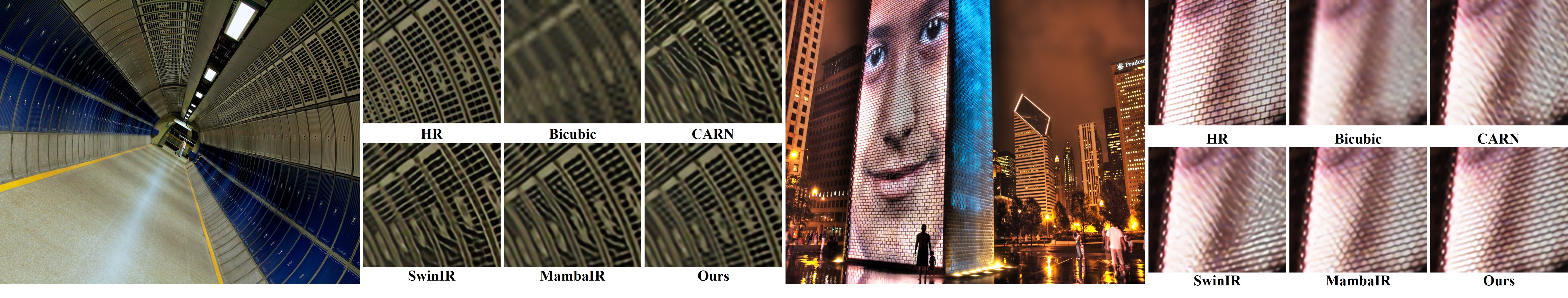}
\captionof{figure}{Quantitative evaluation results on the super-resolution Urban100\cite{urban100} dataset.}\label{fig:super}
\end{minipage}

\begin{minipage}[c]{\textwidth}
\includegraphics[width=\linewidth]{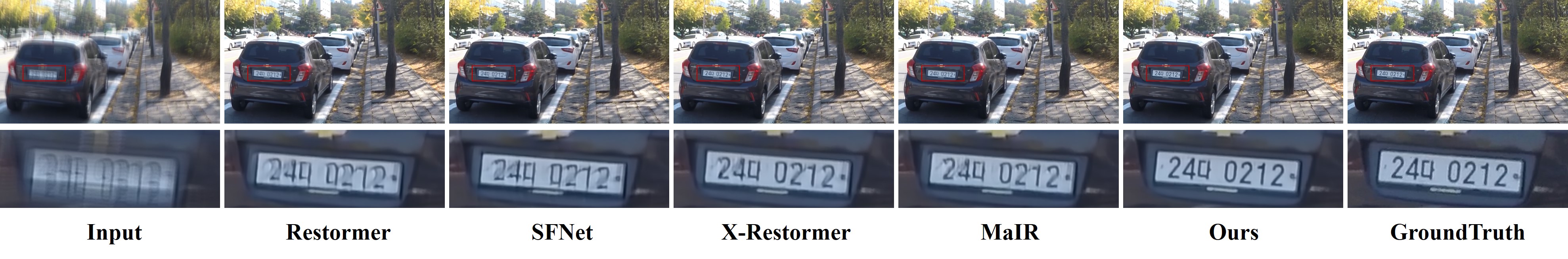}
\captionof{figure}{Quantitative evaluation results on the motion deblurring GoPro\cite{GOPRO} dataset.}\label{fig:blur}
\end{minipage}
\vspace{-3mm}
\end{table*}



\subsection{Experimental results on motion deblurring}
For the motion deblurring task, we selected various state-of-the-art methods for comparison, including CNN-based approaches such as MIMO+\cite{mimo}, MPRNet\cite{MPRNet}, NAFNet\cite{NAFNet} and SFNet\cite{sfnet}, transformer-based models like Restormer\cite{Restormer}, Uformer\cite{Uformer} and X-Restormer\cite{xrestormer}, as well as Mamba-based methods like ACL\cite{acl} and MaIR\cite{mair}. Tab.\ref{tab:blur} presents the quantitative comparison results on the GoPro\cite{GOPRO} dataset. Compared to these state-of-the-art methods, our PW-FNet demonstrates the best performance while achieving an excellent balance between model performance and efficiency. Specifically, Compared to the latest MaIR \cite{mair}, our method outperforms in both PSNR and SSIM, achieving a 0.34 dB PSNR increase with only 22.7\% of the parameters and 25.8\% of the computational cost. Fig.\ref{fig:blur} shows the qualitative comparison results on the GoPro dataset. Our model clearly outperforms others in restoring high-frequency details, particularly in blurred license plate characters, achieving precise and accurate restoration, while other methods still exhibit ghosting artifacts.

\subsection{Experimental results on efficient super-resolution}
For the efficient image super-resolution task, we compare our method with six advanced state-of-the-art approaches, including CNN-based models such as CARN\cite{carn} and ELAN\cite{elan}, transformer-based models like SwinIR-light\cite{swinir} and SRFormer-light\cite{srformer}, and Mamba-based models including MambaIR-light\cite{mambair} and MaIR\cite{mair}. Tab.\ref{tab:super} presents the quantitative comparison results for efficient image super-resolution, with experiments conducted across three different scale settings: $\times$2, $\times$3 and $\times$4. The "Average" column summarizes the performance across five benchmark datasets: Set5\cite{set5}, Set14\cite{set14}, BSD100\cite{bsd100}, Urban100\cite{urban100} and Manga109\cite{manga109}. Compared to other SOTA methods, our PW-FNet consistently achieves superior performance at all three scale settings, while maintaining an efficient balance between model complexity and restoration quality. Specifically, in the more challenging $\times$4 scale setting, our model outperforms MambaIR-light\cite{mambair} with only 69.7\% of the parameters and 41.6\% of the computational complexity, exceeding PSNR by 0.09 dB, 0.16 dB, and 0.28 dB on BSD100, Urban100, Manga109, and 0.16 dB on average, respectively. Fig.\ref{fig:super} illustrates the qualitative comparison results for efficient image super-resolution. The visual results highlight that our method excels in restoring high-frequency details, such as textures and edges, and better preserves the overall consistency of high-frequency components. This performance underscores the effectiveness of our design, which integrates wavelet and Fourier transforms to enable more accurate decomposition and restoration of different frequency bands.



\begin{table*}[t]
\centering
\caption{Quantitative evaluations on dehazing, desnowing, low-light enhancement and underwater enhancement.}\label{tab:other}
\resizebox{\textwidth}{!}{
\begin{tabular}{cccccccccccccc}
\multicolumn{3}{c}{(a) image dehazing}                                                       &  & \multicolumn{3}{c}{(b) image desnowing}                                                      &  & \multicolumn{3}{c}{(c) low-light enhancement}                                                  &  & \multicolumn{2}{c}{(d) underwater enhancement}               \\ \cline{1-3} \cline{5-7} \cline{9-11} \cline{13-14} 
\multicolumn{1}{c|}{\multirow{2}{*}{Methods}} & \multicolumn{1}{c|}{NH-HAZE\cite{nhHAZE}}    & DENSE-HAZE\cite{denseHAZE} &  & \multicolumn{1}{c|}{\multirow{2}{*}{Methods}} & \multicolumn{1}{c|}{SRRS\cite{JSTASRSRRS}}       & Snow100K\cite{desnownesnow100k}   &  & \multicolumn{1}{c|}{\multirow{2}{*}{Methods}} & \multicolumn{1}{c|}{LOL-v2-real\cite{sparselolv2}
} & LOL-v2-syn\cite{sparselolv2}  &  & \multicolumn{1}{c|}{\multirow{2}{*}{Methods}} & LSUI\cite{U-shapeTranslsui}         \\ \cline{2-3} \cline{6-7} \cline{10-11} \cline{14-14} 
\multicolumn{1}{c|}{}                         & \multicolumn{1}{c|}{PSNR/SSIM}  & PSNR/SSIM  &  & \multicolumn{1}{c|}{}                         & \multicolumn{1}{c|}{PSNR/SSIM}  & PSNR/SSIM  &  & \multicolumn{1}{c|}{}                         & \multicolumn{1}{c|}{PSNR/SSIM}   & PSNR/SSIM   &  & \multicolumn{1}{c|}{}                         & PSNR/SSIM    \\ \cline{1-3} \cline{5-7} \cline{9-11} \cline{13-14} 
\multicolumn{1}{c|}{MSBDN\cite{MSBDN}}                    & \multicolumn{1}{c|}{19.23/0.71} & 15.37/0.49 &  & \multicolumn{1}{c|}{DesnowNet\cite{desnownesnow100k}}                & \multicolumn{1}{c|}{20.38/0.84} & 30.50/0.94 &  & \multicolumn{1}{c|}{IPT\cite{ipt}}                      & \multicolumn{1}{c|}{19.80/0.813} & 18.30/0.811 &  & \multicolumn{1}{c|}{UIEWD\cite{uiewd}}                    & 15.43/0.7802 \\
\multicolumn{1}{c|}{FFA-Net\cite{ffaNET}}                  & \multicolumn{1}{c|}{19.87/0.69} & 14.39/0.45 &  & \multicolumn{1}{c|}{CycleGAN\cite{CycleGAN}}                 & \multicolumn{1}{c|}{20.21/0.74} & 26.81/0.89 &  & \multicolumn{1}{c|}{RUAS\cite{RUAS}}                     & \multicolumn{1}{c|}{18.37/0.723} & 16.55/0.652 &  & \multicolumn{1}{c|}{UWCNN\cite{UWCNN}}                    & 18.24/0.8465 \\
\multicolumn{1}{c|}{DeHamer\cite{dehamer}}                  & \multicolumn{1}{c|}{20.66/0.68} & 16.62/0.56 &  & \multicolumn{1}{c|}{JSTASR\cite{JSTASRSRRS}}                   & \multicolumn{1}{c|}{25.82/0.89} & 23.12/0.86 &  & \multicolumn{1}{c|}{Uformer\cite{Uformer}}                  & \multicolumn{1}{c|}{18.82/0.771} & 19.66/0.871 &  & \multicolumn{1}{c|}{Water-Net\cite{U-shapeTranslsui}}                & 19.73/0.8226 \\
\multicolumn{1}{c|}{PMNet\cite{PMnet}}                    & \multicolumn{1}{c|}{20.42/0.73} & 16.79/0.51 &  & \multicolumn{1}{c|}{HDCW-Net\cite{HDCW-NetCSD}}                 & \multicolumn{1}{c|}{27.78/0.92} & 31.54/0.95 &  & \multicolumn{1}{c|}{DRBN\cite{DRBN}}                     & \multicolumn{1}{c|}{20.29/0.831} & 23.22/0.927 &  & \multicolumn{1}{c|}{SCNet\cite{SCNet}}                    & 22.63/0.9176 \\
\multicolumn{1}{c|}{FocalNet\cite{focalnet}}                 & \multicolumn{1}{c|}{20.43/0.79} & 17.07/0.63 &  & \multicolumn{1}{c|}{NAFNet\cite{NAFNet}}                   & \multicolumn{1}{c|}{29.72/0.94} & 32.41/0.95 &  & \multicolumn{1}{c|}{Restormer\cite{Restormer}}                & \multicolumn{1}{c|}{19.94/0.827} & 21.41/0.830 &  & \multicolumn{1}{c|}{U-color\cite{Ucolor}}                  & 22.91/0.8902 \\
\multicolumn{1}{c|}{CPNet\cite{C2PNet}}                    & \multicolumn{1}{c|}{-/-}        & 16.88/0.57 &  & \multicolumn{1}{c|}{FocalNet\cite{focalnet}}                 & \multicolumn{1}{c|}{31.34/0.98} & 33.53/0.95 &  & \multicolumn{1}{c|}{MIRNet\cite{mirnet}}                   & \multicolumn{1}{c|}{20.02/0.820} & 21.94/0.876 &  & \multicolumn{1}{c|}{U-shape\cite{U-shape}}                  & 24.16/0.9322 \\
\multicolumn{1}{c|}{Fourmer\cite{fourmer}}                  & \multicolumn{1}{c|}{19.91/0.72} & 15.95/0.49 &  & \multicolumn{1}{c|}{IRNeXt\cite{IRNeXt}}                   & \multicolumn{1}{c|}{31.91/0.98} & 33.61/0.95 &  & \multicolumn{1}{c|}{SNR-Net\cite{snrnet}}                  & \multicolumn{1}{c|}{21.48/0.849} & 24.14/0.928 &  & \multicolumn{1}{c|}{DM-water\cite{DMwater}}                 & 27.62/0.8867 \\
\multicolumn{1}{c|}{OKNet\cite{OKNet}}                    & \multicolumn{1}{c|}{20.48/0.80} & 16.92/0.64 &  & \multicolumn{1}{c|}{OKNet\cite{OKNet}}                    & \multicolumn{1}{c|}{31.70/0.98} & 33.75/0.95 &  & \multicolumn{1}{c|}{Retinexformer\cite{retinexformer}}            & \multicolumn{1}{c|}{22.80/0.840} & 25.67/0.930 &  & \multicolumn{1}{c|}{WF-Diff\cite{WF-Diff}}                  & 27.26/0.9437 \\ \cline{1-3} \cline{5-7} \cline{9-11} \cline{13-14} 
\multicolumn{1}{c|}{PW-FNet}                  & \multicolumn{1}{c|}{\textbf{20.81/0.82}}           &      \textbf{17.95/0.66}      &  & \multicolumn{1}{c|}{PW-FNet}                  & \multicolumn{1}{c|}{\textbf{32.74/0.98}}           &   \textbf{ 34.50/0.95}        &  & \multicolumn{1}{c|}{PW-FNet}                  & \multicolumn{1}{c|}{\textbf{23.32/0.873}}            &    \textbf{26.02/0.941}         &  & \multicolumn{1}{c|}{PW-FNet}                  &       \textbf{28.44/0.9489}       \\ \cline{1-3} \cline{5-7} \cline{9-11} \cline{13-14} 
\end{tabular}}
\end{table*}

\subsection{Experimental results on other restoration tasks}
Tab.\ref{tab:other} presents the quantitative results of our method on various image restoration tasks, including dehazing, desnowing, low-light enhancement, and underwater enhancement. Compared to recent state-of-the-art approaches, the experimental results demonstrate that our proposed PW-FNet achieves the best performance across all four tasks in terms of PSNR and SSIM. Notably, on the DENSE-HAZE \cite{denseHAZE}, SRRS \cite{JSTASRSRRS}, Snow100K \cite{desnownesnow100k}, and LSUI \cite{U-shapeTranslsui} datasets, our model outperforms the second-best method by an average PSNR margin of 0.8 dB. The substantial performance gains across these diverse degradation scenarios highlight that our proposed PW-FNet holds significant potential to handle a wide range of complex adverse conditions

\subsection{Experimental results on model efficiency}
\begin{table*}[!t]
\centering
\setlength{\tabcolsep}{4pt}
\caption{Quantitative evaluations on model efficiency. Memory and latency is measured on 256$\times$256 resolution images.}\label{tab:infer_com}
\resizebox{\textwidth}{!}{
\begin{tabular}{ccccccccccccccccc}
\cline{1-5} \cline{7-11} \cline{13-17}
\multicolumn{5}{c}{(1)Image Deraining}                                                                             &  & \multicolumn{5}{c}{(2)Motion Deblurring}                                                                          &  & \multicolumn{5}{c}{(3)Image Super-resolution}                                                                      \\ \cline{1-5} \cline{7-11} \cline{13-17} 
\multicolumn{1}{c|}{\multirow{2}{*}{Methods}} & \multicolumn{2}{c|}{Rain200L\cite{RAIN200}}       & \multicolumn{2}{c}{Overhead} &  & \multicolumn{1}{c|}{\multirow{2}{*}{Methods}} & \multicolumn{2}{c|}{GoPro\cite{GOPRO}}         & \multicolumn{2}{c}{Overhead} &  & \multicolumn{1}{c|}{\multirow{2}{*}{Methods}} & \multicolumn{2}{c|}{Urban100\cite{urban100}}       & \multicolumn{2}{c}{Overhead} \\ \cline{2-5} \cline{8-11} \cline{14-17} 
\multicolumn{1}{c|}{}                         & PSNR  & \multicolumn{1}{c|}{SSIM}   & Memory       & Latency       &  & \multicolumn{1}{c|}{}                         & PSNR  & \multicolumn{1}{c|}{SSIM}  & Memory       & Latency       &  & \multicolumn{1}{c|}{}                         & PSNR  & \multicolumn{1}{c|}{SSIM}   & Memory       & Latency       \\ \cline{1-5} \cline{7-11} \cline{13-17} 
\multicolumn{1}{c|}{Restormer\cite{Restormer}}                & 40.99 & \multicolumn{1}{c|}{0.9890} & 668MB        & 85.7ms        &  & \multicolumn{1}{c|}{MPRNet\cite{MPRNet}}                   & 32.66 & \multicolumn{1}{c|}{0.959} & 8959MB       & 1143.2ms      &  & \multicolumn{1}{c|}{ELAN\cite{elan}}                     & 32.76 & \multicolumn{1}{c|}{0.9340} & 258MB        & 45.6ms        \\
\multicolumn{1}{c|}{IDT\cite{IDT}}                      & 40.74 & \multicolumn{1}{c|}{0.9884} & 1285MB       & 79.2ms        &  & \multicolumn{1}{c|}{Restormer\cite{Restormer}}                & 32.92 & \multicolumn{1}{c|}{0.961} & 668MB        & 85.7ms        &  & \multicolumn{1}{c|}{SwinIR-light\cite{swinir}}             & 32.76 & \multicolumn{1}{c|}{0.9340} & 378MB        & 194.0ms       \\
\multicolumn{1}{c|}{DRSformer\cite{DRSformer}}                & 41.23 & \multicolumn{1}{c|}{0.9894} & 1054MB       & 141.9ms       &  & \multicolumn{1}{c|}{SFNet\cite{sfnet}}                    & 33.27 & \multicolumn{1}{c|}{0.963} & 310MB        & 78.1ms        &  & \multicolumn{1}{c|}{SRFormer-light\cite{srformer}}           & 32.91 & \multicolumn{1}{c|}{0.9353} & 346MB        & 286.3ms       \\
\multicolumn{1}{c|}{NeRD-Rain\cite{chen2024NeRD}}                & 41.71 & \multicolumn{1}{c|}{0.9903} & 447MB        & 130.1ms       &  & \multicolumn{1}{c|}{X-Restormer\cite{xrestormer}}              & 33.44 & \multicolumn{1}{c|}{0.946} & 668MB        & 94.4ms        &  & \multicolumn{1}{c|}{MambaIR-light\cite{mambair}}            & 32.85 & \multicolumn{1}{c|}{0.9349} & 597MB        & 197.8ms       \\
\multicolumn{1}{c|}{FADformer\cite{gao2024FADformer}}                & 41.80 & \multicolumn{1}{c|}{0.9906} & 246MB        & 64.1ms        &  & \multicolumn{1}{c|}{MaIR\cite{mair}}                     & 33.69 & \multicolumn{1}{c|}{0.966} & 1400MB       & 996.8ms       &  & \multicolumn{1}{c|}{MaIR\cite{mair}}                     & 32.89 & \multicolumn{1}{c|}{0.9346} & 478MB        & 121.2ms       \\ \cline{1-5} \cline{7-11} \cline{13-17} 
\multicolumn{1}{c|}{(Ours)PW-FNet}            & \textbf{42.23} & \multicolumn{1}{c|}{\textbf{0.9915}} & \textbf{214MB}        & \textbf{27.7ms }       &  & \multicolumn{1}{c|}{(Ours)PW-FNet}            & \textbf{34.03} & \multicolumn{1}{c|}{\textbf{0.968}} & \textbf{259MB}        & \textbf{28.6ms}        &  & \multicolumn{1}{c|}{(Ours)PW-FNet}            & \textbf{33.04} & \multicolumn{1}{c|}{\textbf{0.9369}} & \textbf{215MB}        & \textbf{29.7ms}        \\ \cline{1-5} \cline{7-11} \cline{13-17} 
\end{tabular}}
\vspace{-2mm}
\end{table*}

In addition to model parameter size and computational complexity, we further investigate the inference performance of PW-FNet in comparison with other state-of-the-art models. Tab.\ref{tab:infer_com} presents the quantitative results for deraining, motion deblurring and super-resolution tasks, where we evaluate two key metrics: GPU memory usage and inference latency, both measured at a resolution of 256×256. The experimental results demonstrate that, compared to other advanced methods, our PW-FNet not only achieves the best performance across these tasks but also maintains the lowest memory usage and inference latency, highlighting its potential for practical applications. For instance, in the motion deblurring task, our model outperforms the CNN-based MPRNet\cite{MPRNet} by 1.37 dB in PSNR, while using just 2.9\% of the memory and 2.5\% of the inference time (ours vs. MPRNet: 259MB vs. 8959MB, 28.6ms vs. 1143.2ms). In comparison with the transformer-based X-Restormer\cite{xrestormer}, our model exceeds its PSNR by 0.59 dB, while requiring only 38.8\% of the memory and 30.3\% of the inference time (ours vs. X-Restormer: 259MB vs. 668MB, 28.6ms vs. 94.4ms). When compared to the Mamba-based MaIR\cite{mair}, PW-FNet achieves 0.34 dB higher PSNR, with only 18.5\% of the memory usage and 2.9\% of the inference time (ours vs. MaIR: 259MB vs. 1400MB, 28.6ms vs. 996.8ms). Finally, compared to the frequency filtering-based SFNet\cite{sfnet}, our model surpasses it by 0.76 dB in PSNR, while using just 83.5\% of the memory and 36.6\% of the inference time (ours vs. SFNet: 259MB vs. 310MB, 28.6ms vs. 78.1ms). These results further validate the effectiveness of our approach, where the combination of wavelet-Fourier operations and multi-scale, multi-frequency global modeling enables highly efficient image restoration.

\subsection{Experimental results on real-world generalization}

\begin{table}[t]
\centering
\caption{Quantitative evaluations on real-world deraining.}\label{tab:real_rain}
\resizebox{0.49\textwidth}{!}{
\begin{tabular}{cccccc}
\hline
Methods  & \makecell[c]{IDT \\ \cite{IDT}}       & \makecell[c]{DRSformer\\\cite{DRSformer}}  & \makecell[c]{NeRD-Rain\\\cite{chen2024NeRD}}  & \makecell[c]{FADformer\\\cite{gao2024FADformer}}  & PW-FNet    \\ \hline
\multicolumn{6}{c}{Metric: NIQE$\downarrow$/BRISQUE$\downarrow$}                                  \\ \hline
\makecell[c]{GT-Rain\\\cite{GT-RAIN}}  & 6.34/41.22 & 5.90/38.98 & 5.85/41.08 & 6.54/41.47 &\textbf{5.65/38.77} \\ \hline
\makecell[c]{LHP-Rain\\\cite{lhp-rain}} & 6.84/43.36 & 6.47/43.87 & 6.78/43.66 & 6.79/42.23 &\textbf{6.03/39.06} \\ \hline
\end{tabular}}
\vspace{-3mm}
\end{table}

To evaluate the generalizability of our model in out-of-distribution real-world scenarios, we use a model pre-trained on the SPA-Data\cite{SPA-Data} dataset and directly apply it to two actual rainy condition datasets: GT-Rain\cite{GT-RAIN} and LHP-Rain\cite{lhp-rain}, which contain 2100 and 1000 rainy scenes, respectively. Tab.\ref{tab:real_rain} presents the quantitative comparison results between our method and other deraining baselines. Since real-world deraining scenarios do not have ground truth, we employ two widely used no-reference metrics, NIQE\cite{niqe} and BRISQUE\cite{bri}, to assess the deraining performance. The experimental results demonstrate that, compared to other advanced methods, our PW-FNet achieves the best deraining performance in real rainy scenes.

\subsection{Experimental results on downstream vision task}
One of the core functions of image restoration is to provide high-quality inputs for downstream vision tasks as preprocessing. To evaluate the impact of our proposed model on subsequent tasks, we conduct object detection experiments for autonomous driving using the Rainy in Driving\cite{rainyindriving} dataset. This dataset contains 2496 real-world rainy driving scenes, with a total of 7332 cars, 1135 persons and 613 buses. Tab.\ref{tab:det} presents the quantitative comparison results on the Rainy in Driving dataset, with an IoU threshold set to 0.5. We adopt YOLOv6\cite{li2022yolov6} as the base detection model. Compared to other deraining methods, our model demonstrates superior detection performance across the categories of person, car and bus, meanwhile achieving the best model efficiency. For instance, compared to NeRD-Rain\cite{chen2024NeRD}, our PW-FNet outperforms in the mAP detection metric by 0.01, while utilizing only 3.1\% of the parameters (0.72M vs. 22.89M), 9.4\% of the computational complexity (55.6G vs. 591.9G), 13.3\% of the inference latency (55.7ms vs. 418.8ms), and 36.5\% of the memory usage (0.65GB vs. 1.78GB), with all metrics computed at a resolution of 512×512. Fig.\ref{fig:det} shows the qualitative comparison results for object detection after applying different deraining methods. Our model achieves the best detection outcomes, while IDT\cite{IDT} and DRSformer\cite{DRSformer} exhibit issues with false positives and missed detections.

\begin{table*}[t]
\setlength{\abovecaptionskip}{0.0cm}
\setlength{\belowcaptionskip}{0.0cm}
\centering
\begin{minipage}[c]{\textwidth}
\caption{Quantitative evaluations on downstream detection task.}\label{tab:det}
\centering
\resizebox{0.8\textwidth}{!}{
\begin{tabular}{cccccccc}
\hline
\multicolumn{2}{c|}{Methods}                                                      & Rainy & IDT\cite{IDT}    & DRSformer\cite{DRSformer} & NeRD-Rain\cite{chen2024NeRD} & FADformer\cite{gao2024FADformer} & PW-FNet \\ \hline
\multicolumn{8}{c}{IoU threshold = 0.5}                                                                                                          \\ \hline
\multicolumn{1}{c|}{\multirow{3}{*}{Class(AP$\uparrow$)}}    & \multicolumn{1}{c|}{Person} & 0.274 & 0.280  & 0.284     & 0.290     & 0.276     & \textbf{0.290}   \\
\multicolumn{1}{c|}{}                          & \multicolumn{1}{c|}{Car}    & 0.356 & 0.382  & 0.371     & 0.405     & 0.396     & \textbf{0.421}   \\
\multicolumn{1}{c|}{}                          & \multicolumn{1}{c|}{Bus}    & 0.281 & 0.294  & 0.297     & 0.317     & 0.326     & \textbf{0.330}   \\ \hline
\multicolumn{1}{c|}{Average}                   & \multicolumn{1}{c|}{mAP$\uparrow$}         & 0.304 & 0.319  & 0.317     & 0.337     & 0.333     & \textbf{0.347}   \\ \hline
\multicolumn{1}{c|}{\multirow{4}{*}{Overhead}} & \multicolumn{1}{c|}{\#Params}      & -     & 16.39M & 33.70M    & 22.89M    & 6.96M     & \textbf{0.72M}   \\ \cline{2-8} 
\multicolumn{1}{c|}{}                          & \multicolumn{1}{c|}{\#FLOPs}       & -     & 241.8G & 610.9G    & 591.9G    & 193.3G    & \textbf{55.6G}  \\ \cline{2-8} 
\multicolumn{1}{c|}{}                          & \multicolumn{1}{c|}{Latency}     &   22.7ms    &    476.6ms    &    532.7ms       &     418.8ms      &     169.6ms      &     \textbf{55.7ms}    \\ \cline{2-8} 
\multicolumn{1}{c|}{}                          & \multicolumn{1}{c|}{Memory}      &   0.24GB    &   19.53GB     &      5.33GB     &     1.78GB      &     1.21GB      &   \textbf{0.65GB}      \\ \hline
\end{tabular}}
\vspace{1mm}
\end{minipage}

\begin{minipage}[c]{\textwidth}
\includegraphics[width=\linewidth]{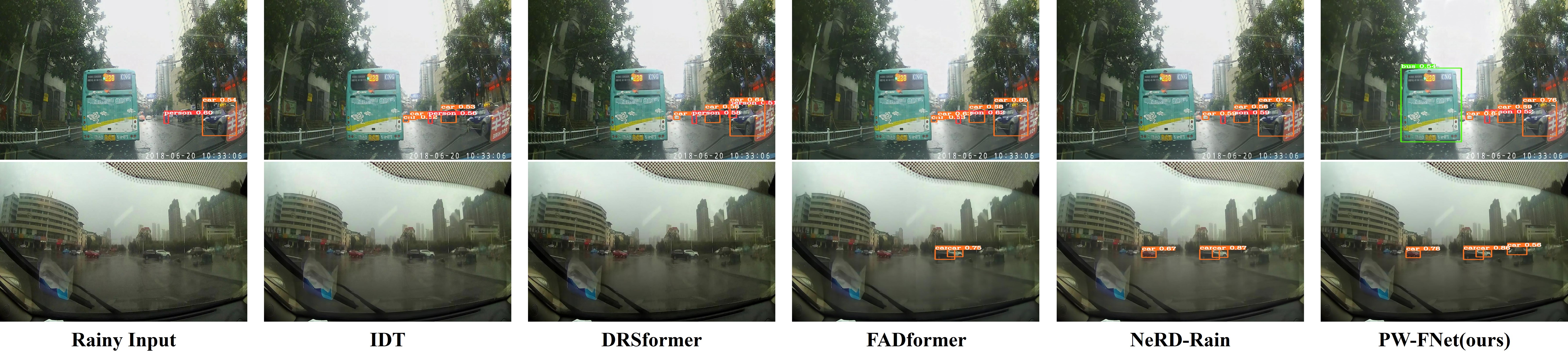}
\captionof{figure}{Quantitative results on the downstream detection Rainy in Driving\cite{rainyindriving} dataset.}\label{fig:det}
\end{minipage}
\vspace{-2mm}
\end{table*}

\subsection{Ablation Study}
To demonstrate the effectiveness of PW-FNet, we conduct ablation studies on Rain200L\cite{yang2017deepRain200} dataset using PSNR/SSIM.

\begin{table}[h]
\setlength{\abovecaptionskip}{0.0cm}
\setlength{\belowcaptionskip}{0.0cm}
\centering
\caption{Ablation on PW-MIMO.}\label{tab:ab_mimo}
\begin{tabular}{c|c|c|c}
\hline
       & SISO   & MIMO   & PW-MIMO \\ \hline
PSNR   & 41.86  & 42.11  & 42.23   \\
\#Params & 1.358M & 1.429M  & 1.442M  \\
FLOPs  & 31.99G & 33.35G & 33.56G  \\ \hline
\end{tabular}
\end{table}

\noindent \textbf{Improvements in PW-MIMO structure.} Tab.\ref{tab:ab_mimo} presents the ablation study on different inter-block structures. PW-MIMO achieves a 0.12 dB PSNR improvement over MIMO at the cost of only 0.013M additional parameters and 0.21G more computational overhead. Compared to SISO, it gains 0.37 dB PSNR while increasing the parameter size by 0.084M and the computational cost by 1.57G.

\begin{table}[h]
\setlength{\abovecaptionskip}{0.0cm}
\setlength{\belowcaptionskip}{0.0cm}
\centering
\caption{Ablation study on choice of wavelet families.}\label{tab:ab_wave}
\resizebox{0.47\textwidth}{!}{
\begin{tabular}{|c|c|c|c|c|}
\hline
Haar         & Daubechies   & Symlets      & Coiflets     & Biorthogonal \\ \hline
42.19/0.9913 & 42.23/0.9915 & 42.27/0.9915 & 42.39/0.9916 & \textbf{42.43/0.9917} \\ \hline
\end{tabular}}
\vspace{-1mm}
\end{table}

\noindent \textbf{Wavelet family choices.} Tab.\ref{tab:ab_wave} illustrates the effects of different wavelet types on model performance, with the Biorthogonal wavelet delivering the best results, while the Haar wavelet shows the least favorable performance. In our work, we select the Daubechies wavelet, which provides an optimal balance between speed and performance.

\begin{table}[h]
\setlength{\abovecaptionskip}{0.0cm}
\setlength{\belowcaptionskip}{0.0cm}
\centering
\caption{Ablation study on Fourier kernel sizes.}\label{tab:ab_fourier}
\resizebox{0.47\textwidth}{!}{
\begin{tabular}{|c|c|c|c|c|}
\hline
8$\times$8 kernel          & 16$\times$16 kernel        & 32$\times$32 kernel        & 64$\times$64 kernel        & Global FFT   \\ \hline
41.79/0.9903 & 41.86/0.9906 & 41.95/0.9909 & 42.08/0.9911 & \textbf{42.23/0.9915} \\ \hline
\end{tabular}}
\vspace{-1mm}
\end{table}

\noindent \textbf{Fourier kernel size choices.} The size of the Fourier kernel directly influences the receptive field range. For instance, in \cite{FFTformer}, window FFT was applied to 8×8 patches rather than the entire image resolution. Tab.\ref{tab:ab_fourier} explores the effect of different Fourier kernel sizes on image restoration performance, revealing that larger Fourier kernel sizes lead to improved restoration outcomes. In our work, we opt for global FFT, applying the transform to the entire image to achieve the maximum global receptive field.

\begin{table}[h]
\setlength{\abovecaptionskip}{0.0cm}
\setlength{\belowcaptionskip}{0.0cm}
\centering
\caption{Ablation study on loss function.}\label{tab:ab_loss}
\resizebox{0.42\textwidth}{!}{
\begin{tabular}{ccc|c}
\hline
Spatial loss & Wavelet loss & Fourier loss & PSNR$\uparrow$/SSIM$\uparrow$    \\ \hline
\checkmark   &              &              & 41.28/0.9901 \\
             & \checkmark   &              & 41.43/0.9904 \\
             &              & \checkmark   & \textbf{42.23/0.9915} \\ \hline
             & \checkmark   & \checkmark   & 42.13/0.9910 \\
\checkmark   &              & \checkmark   & 41.83/0.9909 \\
\checkmark   & \checkmark   &              & 41.38/0.9903 \\ \hline
\checkmark   & \checkmark   & \checkmark   & 41.91/0.9912 \\ \hline
\end{tabular}}
\end{table}

\noindent \textbf{Loss function choices.} As shown in Tab.\ref{tab:ab_loss}, we designed several combinations of loss functions, including spatial loss, wavelet loss and Fourier loss. To our surprise, using Fourier loss alone yielded the best restoration results. This finding aligns with our model design, where the primary feature extraction operations are concentrated in the Fourier domain.

\begin{table}[h]
\setlength{\abovecaptionskip}{0.0cm}
\setlength{\belowcaptionskip}{0.0cm}
\centering
\caption{Wavelet and Fourier-based methods comparison.}\label{tab:ab_wf}
\resizebox{0.49\textwidth}{!}{
\begin{tabular}{cccccc}
\hline
Methods  & DAWN         & DMSR         & WaveMamba    & FADformer    & PW-FNet      \\ \hline
Rain200L & 40.25/0.9879 & 38.81/0.9822 & 39.99/0.9867 & 41.80/0.9906 & \textbf{42.23/0.9915} \\
Rain200H & 31.54/0.9278 &    30.30/0.9089    &  30.81/0.9175  & 32.48/0.9359 & \textbf{32.88/0.9413} \\ \hline
Params   & 3.36M        &     35.42M         &       7.89M       & 6.96M        & \textbf{1.44M}        \\
FLOPs    & 49.61G       &     314.97G     &      55.34G        & 48.51G       & \textbf{33.56G}       \\
Latency  & 102.8ms      &     59.5ms         &        116.0ms      &      69.3ms        &     \textbf{29.4ms}         \\ \hline
\end{tabular}}
\end{table}

\noindent \textbf{Comparison with wavelet-based and Fourier-based methods.} To assess the effectiveness of combining wavelet and Fourier, we compared our approach with two wavelet-based baselines: DAWN\cite{dawn} and WaveMamba\cite{wavemamba}, as well as two Fourier-based baselines: FADformer\cite{gao2024FADformer} and DMSR\cite{DMSR}. Tab.\ref{tab:ab_wf} presents the quantitative comparison results, demonstrating that our proposed baseline, PW-FNet, achieves the best performance with the fewest parameters, lowest computational cost and fastest inference latency, all computed at a resolution of 256×256.




\section{Conclusion}
In this paper, we investigate the significance of global modeling in image restoration and provide an overview of current efficient strategies for global modeling, highlighting that the key to success lies in a model's global modeling capability, with self-attention mechanisms representing just one potential path to achieve this. Building on this perspective, we analyze image degradation patterns within the wavelet-Fourier domain and identify that wavelet-Fourier transforms enable efficient multi-scale, multi-frequency global modeling, offering a viable alternative to self-attention mechanisms. Based on above insights, we introduce a fast, lightweight and effective restoration baseline, PW-FNet, that integrates wavelet-Fourier transforms into both the inter-block and intra-block designs, facilitating efficient image restoration. Extensive experiments on various tasks, including image deraining, raindrop removal, image super-resolution, motion deblurring, image dehazing, image desnowing, and underwater/low-light enhancement, demonstrate that PW-FNet not only outperforms state-of-the-art methods in restoration quality but also achieves superior efficiency, significantly reducing parameter size, computational cost, and inference time.

\bibliographystyle{IEEEtran}
\bibliography{main}

\end{document}